\documentclass{article}

\usepackage{arxiv}

\usepackage[utf8]{inputenc} 
\usepackage[T1]{fontenc}    
\usepackage{amsmath, nccmath, bm}
\usepackage{enumitem}
\usepackage{graphicx,subfigure, multicol}
\usepackage{capt-of}
\usepackage{tabu}
\usepackage{hyperref}       
\usepackage{url}            
\usepackage{booktabs}       
\usepackage{amsfonts}       
\usepackage{nicefrac}       
\usepackage{microtype}      
\usepackage{lipsum}
\usepackage{graphicx}
\graphicspath{ {./} }
\DeclareMathOperator{\E}{\mathbb{E}}
\DeclareUnicodeCharacter{FB01}{\,}
\DeclareUnicodeCharacter{0395}{$E$}
\DeclareUnicodeCharacter{2212}{-}
\title{Robust Topology Optimization Using Multi-Fidelity Variational Autoencoders}

\author{
Rini Jasmine Gladstone \\
  University of Illinois at Urbana-Champaign \\
   \And
 Mohammad Amin Nabian \\
  NVIDIA \\
  \And
 Vahid Keshavarzzadeh \\
  General Motors \\
  \And
 Hadi~Meidani \\
  University of Illinois at Urbana-Champaign\\
  \texttt{meidani@illinois.edu} \\
}

\begin{document}
\maketitle
\begin{abstract}
Robust topology optimization (RTO), as a class of topology optimization problems, identifies a design with the best average performance while reducing the response sensitivity to input uncertainties, e.g. load uncertainty. Solving RTO is computationally challenging as it requires repetitive finite element solutions for different candidate designs and different samples of random inputs. To address this challenge, a neural network method is proposed that offers computational efficiency because (1) it builds and explores a low dimensional search space which is parameterized using deterministically optimal designs corresponding to different realizations of random inputs, and (2) the probabilistic performance measure for each design candidate is predicted by a neural network surrogate. This method bypasses the numerous finite element response evaluations that are needed in the standard RTO approaches and with minimal training can produce optimal designs with better performance measures compared to those observed in the training set. Moreover, a multi-fidelity framework is incorporated to the proposed approach to further improve the computational efficiency. Numerical application of the method is shown on the robust design of  L-bracket structure with single point load as well as multiple point loads.
\end{abstract}


\section{Introduction}
\label{intro}

Topology Optimization is a rapidly expanding research area in various fields such as structural and industrial designs, mechanics, applied mathematics, and computer science~\cite{sigmund2003topology}. Structural topology optimization seeks to determine the best distribution of a material within a prescribed design domain of a structure such that a cost function is minimized and certain performance constraints are satisfied. The pioneering numerical solution to solve this problem is a homogenization approach proposed by~\cite{bendsoe1988generating}. Since then,  other solution approaches have been developed including the density-based approach~\cite{bendsoe1989optimal,mlejnek1992some,zhou1991coc}, level set approach~\cite{allaire2002level,allaire2004structural,wang2003level}, topological derivative method~\cite{sokolowski1999topological}, phase field~\cite{bourdin2003design}, and evolutionary approaches~\cite{xie1993simple}. The most popular approach, however,  is the finite element based topology optimization method called SIMP (Solid Isotropic Material with Penalization for intermediate densities)~\cite{bendsoe1989optimal,rozvany1992generalized}. A detailed description of  SIMP can be found in  Section~\ref{dto}. 

The majority of works on structural topology optimization in the past two decades only consider deterministic performance. However, the performance of structures varies due to  inherent uncertainties~\cite{keshavarzzadeh2017topology}, mainly in loading~\cite{dunning2011introducing,guest2008structural}, boundary conditions~\cite{guo2013robust}, material properties~\cite{chen2010level,guo2009confidence,tootkaboni2012topology} and geometry~\cite{chen2011new,lazarov2012topology}. In order to achieve robust and reliable designs, the effects of these uncertainties should be incorporated in the optimization process, giving rise to two types of topology optimization methods, namely Robust Topology Optimization (RTO) and Reliability Based Topology Optimization (RBTO). RTO  methods attempt to maximize the deterministic performance of the structure while minimizing the sensitivity of the performance with respect to random parameters, resulting in a multi-objective optimization problem, capturing the impact of uncertainties in a qualitative sense. For an overview of robust optimization methods, the readers can refer to~\cite{chen2010level} and \cite{beyer2007robust}. On the other hand, in RBTO, introduced by Olhoff in~\cite{kharmanda2004reliability}, reliability analysis is integrated into the element-based topology optimization method by adding a probabilistic constraint while keeping the objective function deterministic. Reliability-based topology optimization was further developed in recent years by various research groups~\cite{allen2004reliability,frangopol2003life,jung2004reliability}. An overview of RBTO methods can be found in~\cite{mozumder2006investigation} and \cite{valdebenito2010survey}. Some of the recent innovations in solving RBTO problems using deep learning and advanced sampling techniques can be found in~\cite{ates2021two} and  \cite{de2021reliability}. Topology optimization under uncertainties is still an open research area, with difficulties  largely attributed to the high-dimensional property of design space which poses great challenges in uncertainty representation, propagation, and  sensitivity analysis~\cite{chen2010level}. 

We focus, in this paper, on RTO methods and their applications, addressing one of their major challenges, which is the computational expenses attributed to the high dimensionality of  design variables~\cite{abueidda2020topology} and multiple iterations of optimizations~\cite{watts2019simple}. This has led the researchers to study how data-driven machine learning algorithms can be used to address this issue. There has been a surge in research interest to use machine learning, specifically deep learning algorithms, in the area of numerical analysis, to enable faster and accurate evaluation of responses of physical systems by exploiting the underlying governing equations of the problem, observational data, or both~\cite{thuerey2021pbdl, karniadakis2021physics}, and offsetting the computational challenges associated with the numerical methods. In the area of design optimization, the application of deep learning has been explored in various problems~\cite{oh2019deep}, such as topology optimization~\cite{banga20183d,cang2019one,lei2019machine,SosnovikOseledets,yu2019deep}, shape parametrization~\cite{burnap2016estimating,umetani2017exploring}, meta modeling~\cite{farimani2017deep,guo2016convolutional,tompson2017accelerating}, material design~\cite{yang2018microstructural,cang2018improving,cang2017microstructure} and design preference~\cite{burnap2016improving}. 

Among the deep learning methods used for improving the computational efficiency of topology optimization problems, convolutional neural network (CNN) models~\cite{lecun1995convolutional} have been used successfully with limited training data~\cite{gu2018novo}, for a single input variable. \cite{ulu2016data}, further, studied how fully connected neural networks can be used for obtaining optimal designs under varying input variables (e.g. loading magnitudes), by training the model to map the prescribed loading condition to the optimal design. In order to improve the computational time, ~\cite{yu2019deep} proposed a CNN-based encoder and decoder to generate a low-resolution structure from a set of optimal topology designs. This low-resolution structure is then upscaled using conditional Generative Adversarial Networks, thus avoiding finite element iterations. Furthermore, ~\cite{guo2018indirect} performed topology optimization on low dimensional latent space using variational autoencoder and style transfer and ~\cite{cang2019one} used active learning to improve the training of a neural network so that it would result in near optimal topologies. 

Apart from using limited data and training the models on low-resolution structure, parametrization of design candidates using neural networks is another approach to achieve computational efficiency. It is an indirect design representation, where  design variables are mapped to a typically low dimensional space, which could then be explored to find the best topology~\cite{gu2018novo}. Since the parameterized mapping has lower dimension, it reduces the computational time significantly. In this approach, autoencoders~\cite{pinaya2020autoencoders} has been used to parameterize the space of complex geometries~\cite{umetani2017exploring}. However, autoencoders might produce infeasible geometries due to the gaps created in the latent space. This could be addressed using variational autoencoders (VAE)~\cite{kingma2019introduction, wu2022inverse}. ~\cite{burnap2016estimating} showed the possibility of parametrization of the 2D shape of an automobile using VAE. A detailed description of VAE and how it differs from autoencoders is given in Section~\ref{vae}. 

Another approach for better computational efficiency in the training of the surrogates is using  multi-fidelity architectures~\cite{song2022transfer} for neural networks. Multi-fidelity methods use a combination of low- and high- fidelity data to maximize the accuracy of the models, thus minimizing the cost associated with data generation of high-resolution training data and high-dimensional model training. Multi-fidelity design optimization~\cite{chakraborty2017surrogate, shah2015multi, tao2019application} has attracted attention as a strategy for solving complex optimization problems in the past few years. This approach combines low- and high- fidelity models to achieve acceptable accuracy with reduced numerical cost and solve optimization problems. Low-fidelity models can be constructed by using a simplified equation or parameters of the problem, or by linearizing a non-linear equation, like in~\cite{yaji2020multifidelity} or by reducing the dimensionality of the original optimization problem. So far, multi-fidelity optimization approaches have been mainly used for parametric optimization problems with small number of design variables~\cite{alexandrov2001overview,leifsson2010multi,yamazaki2013derivative}. These approaches are not applicable to topology optimization problems due to the high-dimensional nature of the problems and the large number of design variables required for expressing a topology. There have been, thus, very little research on the integration of multi-fidelity design optimization and topology optimization for tackling complex design optimization problems. Moreover, most of the deep learning solution approaches for topology optimization are developed for deterministic problems and few studies have addressed robust topology optimization.

Considering the gaps identified above in research, in the field of robust topology optimization, we investigate in this paper, how deep learning algorithms combined with multi-fidelity approaches can facilitate the solution of RTO problems and shape parametrization. In particular, we focus on the compliance minimization problem, where the optimal design of the structure with loading uncertainty is identified. We have adopted a three stage approach to solve this problem. First, we parameterize the high dimensional geometry of the design candidates using a low dimensional representation obtained by VAEs. Secondly, we accelerate the cost (robust compliance) evaluation step in the optimization iterations by replacing the finite element solver with a fully connected deep neural network. Finally, we use gradient descent algorithm to find the optimal design on the low dimensional representation, minimizing the robust compliance. Furthermore, we present a framework to incorporate the concept of multi-fidelity on the proposed VAE and compliance neural networks to further accelerate the training of these models for solving complex, computationally heavy topology optimization problems. Here, training samples for training the neural network models are generated using SIMP or power law approach. 

Thus, the remainder of this paper is organized as follows. A theoretical background on RTO and VAEs is presented in Section 2. Our proposed methodology for the robust topology optimization problem is then introduced in Section 3. Section 4 evaluates the performance of the proposed methodology when applied on a compliance minimization problem. Finally, Section 5 concludes the paper.

\section{Background}
\label{sec:background} 

\subsection{Deterministic Topology Optimization}
\label{dto}
Structural Topology Optimization can be defined as the arrangement of materials within a design domain to optimize the mechanical performance of the structure while accounting for design constraints~\cite{de2020topology}. In this paper, we restrict the discussion of the topology optimization problem to the well known compliance minimization design problem. The goal of this problem is to identify the topologies that store minimum strain energy under a set of applied loads, given a limited volume of material. ~\cite{bendsoe1989optimal}, ~\cite{zhou1991coc}, and ~\cite{mlejnek1992some} proposed power-law or SIMP approach, where the design domain is discretized into elements called isotropic solid micro structures and the properties within these elements are assumed constant. According to ~\cite{bendsoe1989optimal}, topology optimization involves determining, at an element level, whether there is material present or not. Thus, the density distribution of the material in the design domain, \(\bm \theta\), is discretized into these elements, and the density of each element, \(\theta_\text{e}\), is either 0 or 1 depending on whether material is present or not. However, in order to make the density distribution continuous, this density can vary from \(\theta_\text{min}\) and 1, thus allowing intermediate densities for the elements. \(\theta_\text{min}\) is the minimum allowable density for the elements ensuring the numerical stability of finite element analysis. For each element \(e\), the density, \(\theta_\text{e}\), and Young modulus of elasticity of the material, \(E_\text{0}\), is related as \(E(\theta_\text{e})=\theta_e^\beta E_\text{0}\), i.e. the power law. The penalty factor, $\beta$, assigns lower weight to the elements with intermediate densities, thus ensuring that the optimal topology moves towards solid black (\(\theta_\text{e} = 1\)) or void white (\(\theta_\text{e}= \theta_\text{min}\)) design. 

~\cite{sigmund200199} provides a 99 line topology optimization code in MATLAB using the SIMP approach. Here, the design domain is considered to be rectangular and discretized by square finite elements with  \(n_x\) and \(n_y\) being the number of elements in the horizontal and vertical directions. The topology optimization problem to minimize compliance using SIMP approach is written as

\begin{equation}
\begin{aligned}
& \underset{\bm \theta}{\text{minimize}}
& & \mathrm Q(\bm \theta)=U(\bm \theta)^TK(\bm \theta)U(\bm \theta)  \\
& \text{subject to}
& & \frac{V(\bm \theta)}{V_\text{0}}=\alpha_\text{v} \\
&&& K(\bm \theta) U(\bm \theta)=F_\text{0}\\
&&& 0 < \bm \theta_\text{min} \leq \bm \theta \leq 1,
\label{eq:1}
\end{aligned}
\end{equation}
where \(U\) and \(F_\text{0}\) are the global displacement and force vectors, respectively, and \(K\) is the global stiffness matrix. \(\bm \theta_\text{min}\) is a vector of minimum densities, \(V(\bm \theta)\) and \(V_\text{0}\) is the material volume and volume of the design domain, respectively, and \(\alpha_\text{v}\) is the prescribed volume fraction with $ 0 <  \alpha_\text{v} \leq 1$.

The topology optimization problem described above is a deterministic one, as it aims to obtain the solution without taking into account the effects of uncertainties of input variables. The parameters or conditions that could be in general uncertain include the geometry, loading conditions, material properties, etc. In the next section, we present the formulation of a RTO problem which takes into account the impacts of these uncertainties.

\subsection{Robust Topology Optimization}

In RTO problems, a design is identified by minimizing a probabilistic form of the compliance. According to the original objective of RTO, we seek to solve a a two-objective optimization model which simultaneously minimizes the mean of the compliance and also its standard deviation.   An efficient way to solve RTO  is to combine the two objective functions into a single weighted function. For the sake of brevity, and in accordance with the examples in this paper, let us  consider the RTO problem with uncertainty only in the loading. In particular, let $\bm \theta$ be the design variable and $\bm \xi$ be the  random load. Then, the robust topology optimization formulation for minimizing the robust compliance, $Q_\text{rob}(\bm \theta)$ can be written as~\cite{zhifang2016robust}

\begin{equation}
\begin{aligned}
\underset{\bm \theta}{\text{minimize}}  \ \
& \mathrm Q_\text{rob}(\bm \theta)=&& \E_{\bm \xi}  \left[ U(\bm \theta,\bm \xi)^TK(\bm \theta)U(\bm \theta, \bm \xi)\right]+\lambda \  \left(\text{Var}_{\bm \xi} \left[ U(\bm \theta,\bm \xi)^TK(\bm \theta)U(\bm \theta, \bm \xi))\right] \right)^{1/2} \\
& \text{subject to} && \frac{V(\bm \theta)}{V_\text{0}} = \alpha_\text{v}  \\
&&& 0 < \bm \theta_\text{min} \leq \bm \theta \leq 1,
\label{eq.RTO}
\end{aligned}
\end{equation}where $\E_{\bm \xi}$ and $\text{Var}_{\bm \xi}$ are the expectation and variance operators with respect to the loading uncertainty $\bm \xi$.
To calculate the compliance, the displacement for each realization of the random loading is calculated (predominantly using the finite element approach) by solving 
$$
K(\bm \theta) \ U(\bm \theta, \bm \xi)=F_\text{0}(\bm \xi),     \ \ \bm \xi \in S_{\bm \xi},
$$
where $S_{\bm \xi}$ denotes the state space of random inputs. The constant $\lambda$ determines the balance between the mean and standard deviation; $\alpha_{\text{v}}$ is the prescribed volume fraction. 

Robust topology optimization using finite element solvers are computationally expensive for high dimensional designs. In this study, we develop feed-forward fully-connected deep neural network surrogates to accelerate the response evaluations and VAEs to offer a low-dimensional representation for design variables $\bm \theta$.

\subsection{Feed-forward fully-connected deep neural networks}
Feed forward fully connected neural networks, or multi layer perceptrons (MLPs), are deep learning models which serve as a nonlinear mapping \(\bm y=\mathcal{N}(\bm x,\bm W) \), that approximates a target unknown function $\mathcal{N}^*(\bm x)$. The model is built by learning the value of the neural network parameters, \( \bm W\), that results in the best function approximation.  A fully connected neural network consists of a series of fully connected layers. A fully connected layer is a function from \(\mathbb{R}^\text{m}\) to \(\mathbb{R}^\text{n}\), and  \(\bm x \in \mathbb{R}^\text{m}\) represents the input  and \(\bm y \in \mathbb{R}^\text{n}\) is the $n$-dimensional model output. For instance, the output of a single hidden layer neural network, with \(l\) nodes in the layer~\cite{nabian2018deep}, can be calculated as
 
\begin{equation}
\begin{aligned}
 \mathbf{y}=\mathcal{A}( \mathbf{xW}_\text{1}+ \mathbf{b_\text{1}}) \mathbf{W_\text{2}}+ \mathbf{b_\text{2}},
\label{eq:5}
\end{aligned}
\end{equation}
where \(\mathcal{A}(.)\) is a non-linear function, known as the activation function, \(\mathbf{W_\text{1}}\) and \(\mathbf{W_\text{2}}\) are the weight matrices of size \(m \times l\) and \(l \times n\) and \( \mathbf{b_\text{1}}\) and \( \mathbf{b_\text{2}}\) are the bias vectors of size \(1 \times l\) and \(1 \times n\), respectively. A deep neural network has multiple hidden layers, where the output from the activation function from one hidden layer is transformed by new weight and bias values and fed to the next hidden layer~\cite{nabian2018deep}. A non-linear activation function enables the neural network to generate non-linear mappings from inputs to outputs. The most popular activation functions are sigmoid, tanh (Hyperbolic tangent) and ReLU (Rectified Linear Units). The ReLU activation function, one of the most widely used functions, has the form, \(\mathcal{A}(x)=\max(0,x)\). Sigmoid function has the form, \(\mathcal{A}(x)={1}/({1+\exp(-x)})\), and ranges between 0 and 1. The neural network architectures in the present study use ReLU to activate the hidden layers. The output layer of the Variational Autoencoder is activated by sigmoid function to ensure that the density values range between 0 and 1.

\subsection{Variational Autoencoders}
\label{vae}
An autoencoder is a  neural network model that learns a compact representation of a data (e.g. an image or a vector), while retaining its most important features. It consists of a pair of  connected neural networks - an encoder and a decoder. The encoder network takes an input from a typically high-dimensional space and converts it into a lower dimensional representation known as the latent space. The decoder network takes a point from the latent space  and reconstructs a corresponding output in the original space. The encoder and decoder  neural networks are trained together and optimized via back-propagation~\cite{wang2016auto}. In the context of structural topology, the training set includes $n$   topologies, \( [\bm \theta_1,\bm \theta_2,\bm \theta_3,...,\bm \theta_n] \). Using these samples, the training of the autoencoder is carried out by minimizing the reconstruction error,

\begin{equation}
L_{\text{reco}}= \displaystyle\sum_{i=1}^{n} \|\bm  \theta_i- \hat{\bm \theta}_i\|^{2}, \label{eq:6}
\end{equation}

where \(\bm \theta_i\) is the $i$-th topology sample and  \({\hat{\bm \theta}}_i\) is its corresponding reconstructed topology, respectively. The hidden layers could be convolutional, fully connected or could take any other form depending on the task at hand. In this work, we have used fully connected deep neural networks for the encoder and decoder networks. 

The fundamental problem with autoencoders is that the trained latent space may not be continuous, or may not allow easy interpolation~\cite{goodfellow2016deep}. If the latent space has discontinuities (e.g. gaps between clusters) and the topology that is generated by a sample from one of the gaps may be an unrealistic geometry. This is because, the autoencoder, during training, has never seen encoded vectors  from that region of latent space. This challenge could be addressed by the use of Variational Autoencoder (VAE).

A VAE, like an autoencoder, consists of an encoder, a decoder and a loss function. The encoder neural network, denoted by $f(\cdot)$, encodes an input topology, \(\bm{\theta}\), to a (random) latent variable $\bm{z}$, i.e. \(\bm{z} = f(\bm{\theta}) \sim q(\bm{z}|\bm{\theta})\). The decoder network of VAE, denoted by $g(\cdot)$, decodes a given (random)  latent variable, \(\bm{z}\), back to an image similar to the original input, i.e. \(\bm{\hat{\theta}} = g(\bm{z}) \sim p(\bm{\theta}|\bm{z})\). Here, $\bm{\hat{\theta}}$ is the vector of reconstructed images by the decoder network, generated from the latent vector, \(\bm{z}\). Let the latent representation, \(\bm{z}\), be sampled from the prior distribution, \(p(\bm{z})\). The latent space is regularized for the purpose of continuity (two close points in latent space gives similar output) and completeness (the generated output should not be non-meaningful). To satisfy this, it is required to regularize both the covariance matrix and the mean of the latent space distribution~\cite{doersch2016tutorial}. This is done by enforcing the prior distribution to be a standard normal distribution, i.e., $p(\bm{z}) \sim \mathcal{N}(0,1)$. 

The loss function of VAE consists of a reconstruction loss term  and  a regularizer. Let \([\bm \theta_1,\bm \theta_2,\bm \theta_3,...,\bm \theta_n]  \), be the set of topology samples, then the total VAE loss is given by
\begin{equation}
L_{\text{VAE}} = \sum_{i=1}^{n} l_i
\end{equation}
where $l_i$ is the loss function for the $i$th  data point (sampled topology), given by
\begin{equation}
\begin{aligned}
l_i &= -\E_{\bm z \sim q(\bm z \mid \bm \theta_i)}[\log (p(\bm{\theta_i}\mid \bm{z}))] + \text{KL}(q(\bm z\mid \bm \theta_i)\mid \mid p(\bm {z})).
\label{eq.vae}
\end{aligned}
\end{equation}
The first term is the reconstruction loss, or expected negative log-likelihood of the $i$th data point. It measures how likely it is that the data point $\bm \theta_i$ is explained by the random variable $\bm z$.  The  prior distribution for the VAE latent variables, $p(\bm z)$, is taken to be a standard normal distribution. The expectation in Eq.~(\ref{eq.vae}) is taken with respect to $q(\bm z \mid \bm \theta_i)$, which is referred to as the `encoder distribution'. In calculating the likelihood, we use this distribution instead of the prior $p(\bm z)$ (which doesn't depend on $\bm \theta$) in order to promote sampling from $\bm z$ values that are expected to have contribution to the training data. Otherwise, one may obtain many samples from the latent space for which the training data is improbable, i.e. $p(\bm \theta \mid \bm z) = 0$.

Since we use encoder distribution instead of the prior $p(\bm z)$, in the derivation of the loss function, we will also obtain the second term in the left hand side of Eq.~(\ref{eq.vae}), which serves as a regularizer. This term is the Kullback-Leibler (KL) divergence between the encoder distribution $q(\bm z \mid \bm \theta)$ and the prior $p(\bm z)$ and measures how much information is lost when distribution $q$ is instead of $p$. A detailed derivation of the loss function of VAE can be found in~\cite{doersch2016tutorial} and \cite{kingma2019introduction}.

\begin{figure}
\centering
\includegraphics[width=\textwidth]{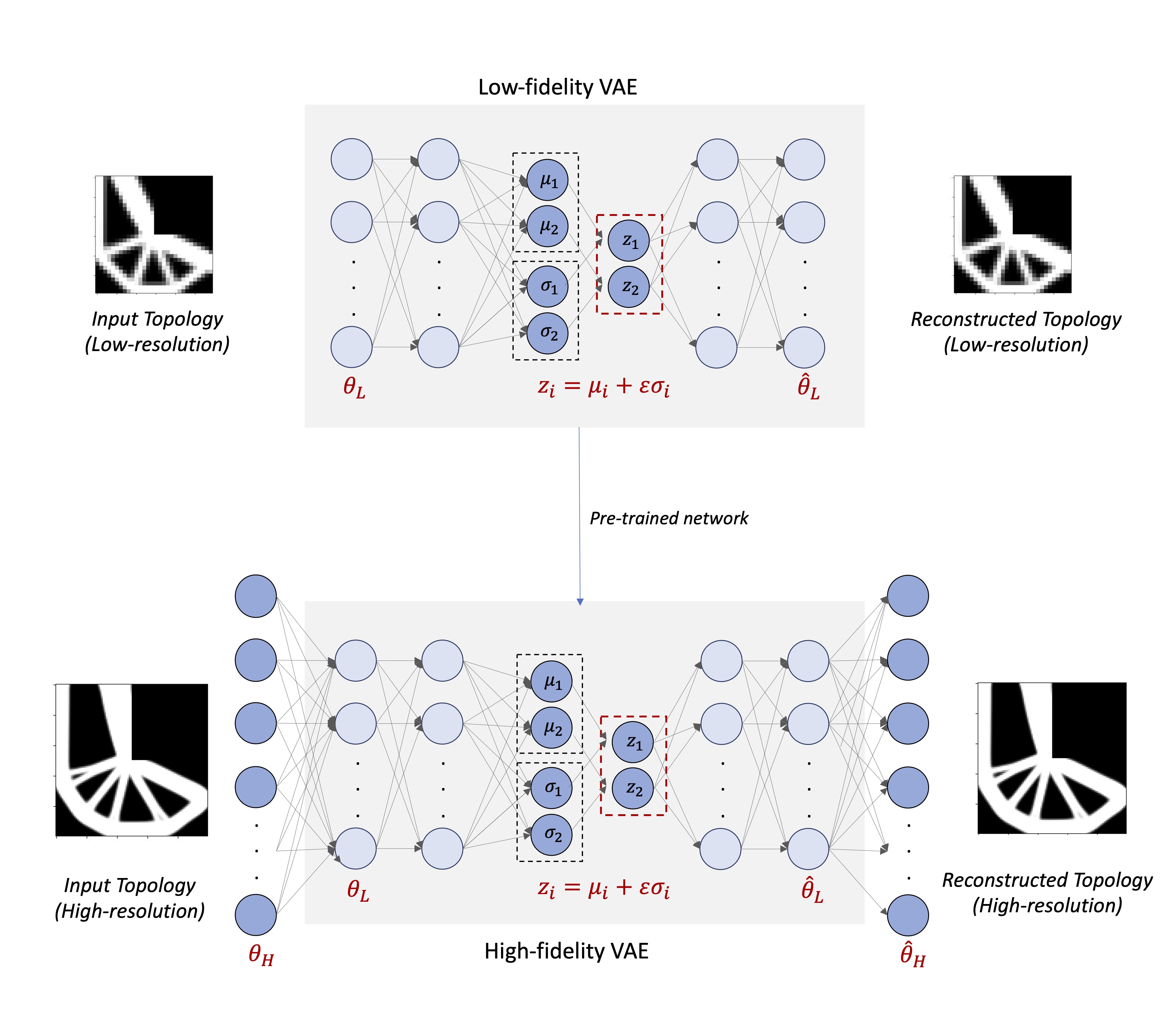}
\caption{An overview of the neural network structure of a VAE with a two dimensional latent space. It consists of an encoder network and decoder network. The input of the encoder network is the image of a topology and the output of the decoder network is the corresponding reconstructed topology. The encoder maps the high dimensional input data to a low dimensional vector, $[z_1, z_2]$, which becomes the input to the decoder network.}
\label{fig:vaeoverview}
\end{figure}

During the training of VAE, the decoder randomly samples the latent vector, \(\mathbf{z}\), from the encoder distribution, \(\mathbf{z} \sim q(\mathbf{z})\). But, the back propagation cannot flow through random sampling and hence, a reparametrization technique is used, where \(\mathbf{z}\) is approximated by the following normal distribution,

\begin{equation}
\begin{aligned} 
\bm{z} \sim \mathcal{N}(\bm \mu, \text{diag}(\bm \sigma)) \label{eq:13}
\end{aligned}
\end{equation}where \(\bm \mu\) and \(\bm \sigma\) are set to be the output vectors from the encoder network, $f$. An overview of the VAE neural network architecture is shown in Fig.~\ref{fig:vaeoverview}.

In the present study, the encoder network consists of input layer, which is the vectorized version of the input topology images, a number of fully connected hidden layers and the latent space layer as the output layer.  The number of nodes in the latent space layer depends on the latent space dimension, $|\bm z|$. There are two outputs from the encoder network - vectors of \(\bm \mu\) and \(\bm \sigma\) values, which are the input to the decoder network. The decoder network, like the encoder, consists of an input layer, with size $|\bm z|$, and the output layer which  gives the reconstructed/generated topology, and thus has the same number of nodes as the input layer of the encoder network. In this work, we use a train VAE as a means to map high dimensional topology data into a low dimensional space, and also map new samples in the latent space into newly generated topologies.

\section {METHODOLOGY}
Using the formulation in Eq.~\ref{eq.RTO}, the objective of the study is to find the optimal design variable, \(\bm \theta^*\), which minimizes \(Q_\text{rob}(\bm \theta)\). We do so by solving the optimization problem minimizing a sample-based approximation of the robust compliance given by
\begin{equation}
\begin{aligned}
\underset{\bm \theta}{\text{minimize}}  \ \
& \mathrm Q_\text{rob}(\bm \theta) \approx && \frac{1}{n} \sum_{i=1}^{n} Q_i(\bm \theta)  + \lambda \left({\frac{1}{n} \sum_{i=1}^{n} (Q_i(\bm \theta)-\mu_\text{Q})^2}\right)^{1/2} \\
& \text{subject to} && \frac{V(\bm \theta)}{V_\text{0}} = \alpha_\text{v}  \\
&&& 0 < \bm \theta_\text{min} \leq \bm \theta \leq 1,
\label{eq:sampleRTO}
\end{aligned}
\end{equation}
where $Q_i(\bm \theta) = U(\bm \theta, \bm \xi_i)^TK(\bm \theta)U(\bm \theta, \bm \xi_i)$,  is the deterministic compliance, calculated based on the  global stiffness matrix $K(\bm \theta)$ and the global displacement vectors for the $i^{\text{th}}$ realization of the random input parameter; and $\mu_\text{Q} := \frac{1}{n} \sum_{i=1}^{n} Q_i(\bm \theta)$. 

The search space for the solution of this optimization problem is the set of all the designs $\bm \theta$, each evaluated at all the realizations of the random parameter, $\bm \xi$. Because of the high dimensional representation of topologies, it is computationally  challenging to carry out optimization in this search space. As a remedy, we consider a subspace of the search space, which consists of only the deterministic designs that are optimal with respect to  particular   realizations of  $\bm \xi$. To create this space, we undertake a supervised learning approach, and  obtain these deterministic designs from the SIMP algorithm given various realizations of input. This search subspace ensures that we have a relatively good initial space of candidate designs, without having to run any RTO solver. 

\begin{figure}
\centering
\includegraphics[width=\textwidth]{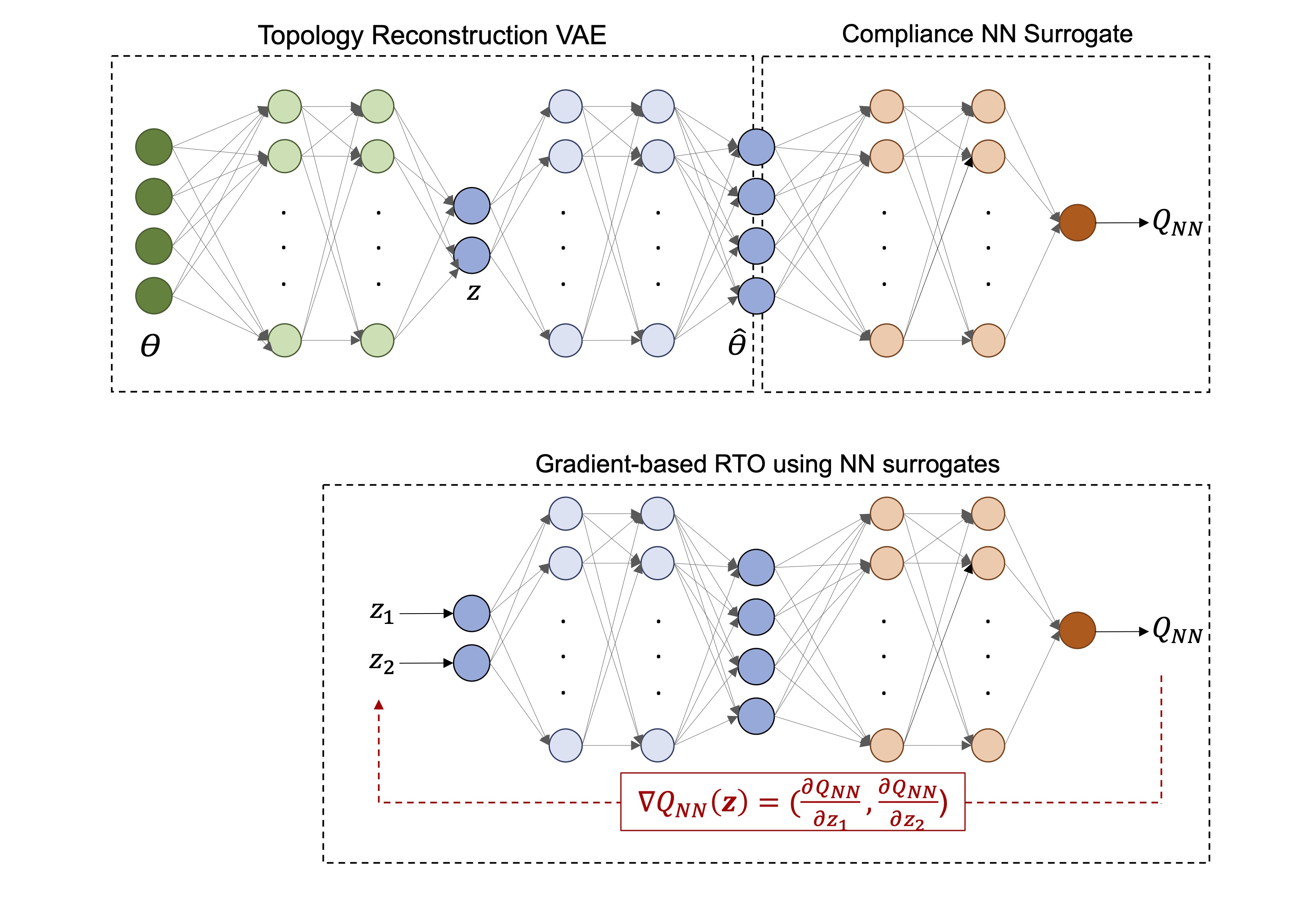}
\caption{The complete neural network architecture for  topology optimization. The top image shows the training of the VAE network and the compliance neural network surrogate. The bottom image shows how gradient descent is applied for finding the optimal design for the compliance minimization problem.}
\label{fig:nnarch}
\end{figure}

In particular, our RTO solution approach, shown in Fig.~\ref{fig:nnarch}, consists of three stages. The first stage is the parameterization of our search subspace to reduce the dimensionality of the design representations. There are a variety of approaches for doing this, a review of which could be found in~\cite{salunke2014airfoil}. In the present study, VAEs are used to represent the geometries in terms of latent space distribution, which could be then used to reconstruct the original designs. One of the main advantages of using VAE is the reduction in dimensionality, from thousands to as low as 2. The second stage is the fast estimation of the robust compliance for candidate designs using a feed forward fully connected compliance neural network surrogate. The final stage is finding the optimal design which minimizes the robust compliance using a gradient-based optimization process, which involves back-propagation through the compliance network and the decoder part of the VAE, as shown in Fig.~\ref{fig:nnarch}. In what follows, we provide details about these networks.

\subsection{Neural Networks for Dimensionality Reduction and Robust Compliance Prediction}

As illustrated in Fig.~\ref{fig:nnarch}, the first part of the solution approach is the dimensionality reduction of the designs using VAE. These designs are the input, $\bm \theta$, to the encoder network of VAE, $f(\cdot)$. The images used for training VAE are a sample of geometries from the search subspace, which are the deterministic optima obtained at various realizations of $\bm \xi$.
The encoder network consists of an input layer, a number of fully connected hidden layers and an output layer, which are the mean and standard deviation values for the latent variable $\bm z$. ReLU function is used as the activation function for all the hidden layers. The decoder network, $D$, consists of an input layer, a number of fully connected hidden layers, an AVG (average) pooling layer and an output layer. The input layer has \(|\bm z|\) number of nodes. ReLU function is used as the activation function for all the hidden layers. 

The output of the last hidden layer, which has the same dimension as the input image,$\bm \theta$, is passed through an AVG pooling layer. The pooling layer acts as a design filter, to ensure that the reconstructed image does not have checker board pattern. It does this by replacing the pixel value with the average of its neighboring pixel values. The output of the AVG pooling layer is, then, passed through a sigmoid function to make sure all the pixel values are between 0 and 1 to get the final reconstructed image. Finally, the entire VAE network is trained by updating its weights and biases to reduce the loss function of Eq.~(\ref{eq.vae}). 

The second part of the solution is to have a computationally efficient method to calculate the robust compliance of the candidate designs, \(Q_\text{rob}(\bm \theta)\). This is done by training a compliance neural network surrogate, which is a feed forward fully connected deep neural network, consisting of an input layer, a number hidden layers and an output layer. Input layer has number of nodes equal to the dimension of the training image. Output layer is a single node which gives predicted robust compliance, \(Q_\text{NN}(\bm \theta)\). The loss function for the training of the network is the mean squared error between \(Q_\text{rob}(\bm \theta)\) and the output of the neural network, \(Q_\text{NN}(\bm \theta)\). ReLU function is used as the activation function for all the layers and Adam optimizer is used for the training of the neural network with a constant learning rate.

\subsection{Gradient-based RTO via Low Dimensional Space}
\label{GD}

The final stage of the solution is to find the optimal design from the generated designs from VAE. One of the fundamental methods is the brute force method, where we generate various designs from the search space, compute \(Q_\text{NN}\) using compliance neural network surrogate and find the design with the minimum compliance. However, this is not an efficient approach and would not guarantee that the design is the most optimal one. Therefore, we use a gradient-descent algorithm for finding the most optimal design for the topology optimization problem.

Using stochastic gradient descent on on the compliance surrogate, one can optimize the topology according to the robust compliance criterion. Specifically, at step \(i+1\) of the gradient descent, we can use the following update rule
$$
{\bm \theta}_\text{i+1}={\bm \theta}_\text{i}-\eta \nabla {Q_\text{NN}}(\bm \theta_\text{i}),
$$
where ${\bm \theta}_\text{i}$ and ${\bm \theta}_\text{i+1}$ are the designs at steps $i$ and $i+1$ of the gradient descent journey, and $\nabla {Q_\text{NN}}(\bm \theta_\text{i})$ is the gradient of the compliance calculated on the $i$th geometry. 

This update rule is not efficient, however, as the topology \(\bm \theta\) is typically high dimensional and this can significantly slow down the algorithm. To improve the computational speed and reduce the dimensionality of the input, we instead consider the lower network in Fig.~\ref{fig:nnarch}, which consists of the decoder part of the VAE and the compliance surrogate. Using this network, we compute the gradient of the robust compliance with respect to the latent space encoding of the geometry, i.e. \(\bm z\). Therefore, the gradient descent update can be rewritten as
$$
{\bm z}_\text{i+1}={\bm z}_\text{i}-\eta \nabla {Q_\text{NN}}(\bm z_\text{i}),
$$
and the dimension of the gradient calculation is reduced to \(|\bm z|\). Using this update rule, we effectively search the lower dimensional latent space for the optimal $\bm z$. The gradient descent updates begin on a randomly selected initial point in the latent space. It is imperative to try multiple starting points to ensure that the algorithm does not converge to a local minima with a poor compliance.

So we effectively identify the robust design by solving the following unconstrained optimization 
\begin{equation}
\begin{aligned}
\underset{\bm z}{\text{minimize}} \ \  \mathrm Q_\text{NN}(\bm z)= h \circ g  \ (\bm z),
\label{eq.zRTO}
\end{aligned}
\end{equation}
where $g(\cdot)$ is the decoder part of the VAE which serves as the generator of candidate designs, and $h(\cdot)$ is the approximate predictive model for robust compliance.

\subsection{Multi-fidelity VAE and compliance NN surrogates}

\begin{figure} 
\centering
\includegraphics[width=\textwidth]{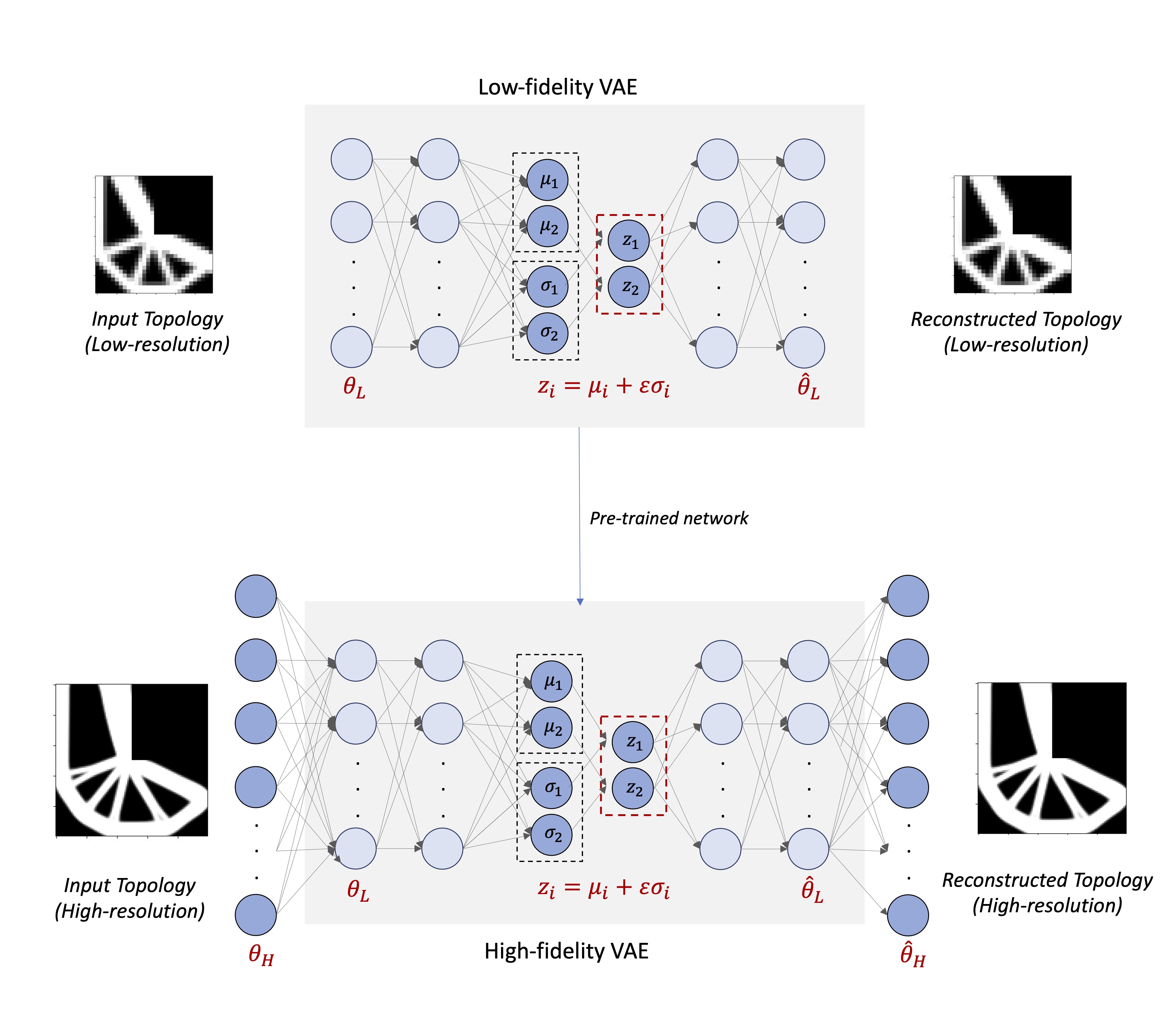}
\caption{An overview of the VAE architecture with multi-fidelity approach. A low-fidelity VAE network is trained using a sample of low-resolution designs, $\theta_L$. This is used as a pre-trained network to which additional fully connected layers are added to the encoder and decoder parts to form the high-fidelity VAE. The additional layers are added to match the dimensions of high-resolution designs, $\theta_H$, where $|\theta_H|>>|\theta_L|$.}
\label{fig:vaeoverview_mf}
\end{figure}

Even as the proposed framework addresses the high-dimensional nature of the topology optimization problem by identifying a low-dimensional latent space, one still needs a large number of high-dimensional topology samples for  training. This can be time  consuming and pose a computational burden. To address this challenge, we propose a multi-fidelity training  approach, shown in Fig.~\ref{fig:vaeoverview_mf} in particular for the VAE training.  The framework consists of a low-fidelity VAE, which uses low-resolution images, $\theta_L$, and their corresponding reconstructed images with the same low-resolution. Since the low-fidelity VAE is built with fewer parameters, and the training set consists of cheap low-resolution samples, the data generation and  training is significantly faster. 

Once the low-fidelity VAE is trained, we construct a high-fidelity VAE by adding one or two layers to the first and last layers of the low-fidelity VAE. The input and output dimensions are equal to the dimension of high-resolution topologies. It can be noted that the resulting multi-fidelity VAE may be designed with the same size as a single-fidelity VAE with high-resolution topologies (e.g. the one shown in Fig.~\ref{fig:vaeoverview}), but the advantage is the core of the multifidelity VAE consists of a pre-trained low-fidelity model. This means that one would require only a fraction of the high-resolution training data compared to the single-fidelity VAE.

A similar multi-fidelity training can be considered for the compliance NN surrogate  as well, where a smaller fully connected low-fidelity compliance neural network is trained to serve as the pre-trained part of a larger high-fidelity compliance network. The low-fidelity compliance neural network surrogate takes a large number of low-resolution designs, $\theta_L$ as the input and predicts their robust compliance values, $Q_L$. The high-fidelity compliance neural network surrogate is, then, constructed by adding one or two layers before the first layer of the low-fidelity surrogate. This ensures that the first layer of the high-fidelity surrogate matches the dimensions of the high-fidelity designs, $\theta_H$. Similar to VAE, the multi-fidelity compliance surrogate model has the same size as single-fidelity compliance network trained with just high-resolution topologies. The trained parameters from the low-fidelity compliance neural network surrogate are used as the initialized parameter values for the training of high-fidelity compliance neural network surrogate. This means that the training of the high-fidelity model requires only a small number of high-resolution designs.

\section {RESULTS}
In this section, we demonstrate the effectiveness of our proposed methodology in solving the robust topology optimization problem with single load application for an L-shaped bracket with uncertainty in load angle. We also carry out robust topology optimization using the proposed approach for L-shaped bracket with multiple load uncertainty. For both of these examples, we show the effectiveness of using multi-fidelity approaches to the network architecture in terms of improvement in the computational efficiency and finding the optimal designs.

\subsection{L-bracket with single loading uncertainty}
\begin{figure} 
\centering
\includegraphics[width=0.6\textwidth]{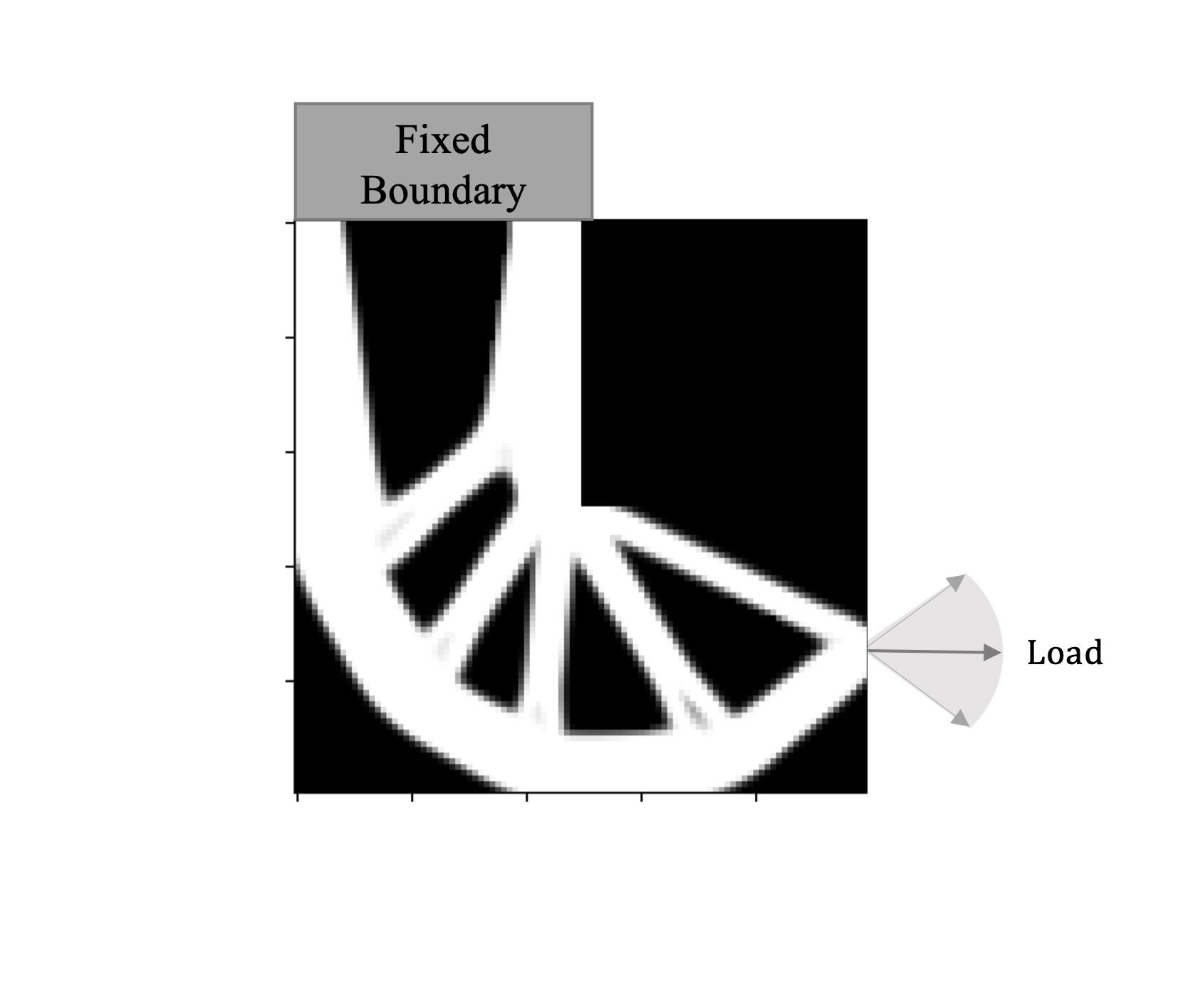}
\caption{An example of the structure used in our study. The top part of the L-shaped structure has a fixed boundary. The load is applied at a single point with uncertainty in the loading angle. }
\label{fig:geoeg}
\end{figure}

As the first example, we  consider an L-shaped structure with fixed boundary and a single point load. The applied load has a fixed magnitude but a known random angle. An example of a candidate topology is shown in Fig.~\ref{fig:geoeg}. The load angle is calculated as the angle from the downward vertical line in the counter clockwise direction and is considered to be uniformly distributed between 0 and $\pi$. In this example, the resolution of the images representing the topology is considered to be $100\times100$, and therefore the input \(\bm \theta\),  defining the geometry of the structure, is a \(n_{\bm \theta}=10,000\) dimensional vector. The random input \(\bm \xi\) in this case is a single random variable (i.e. \(n_{\xi}=1\)), namely the load angle. The objective is to find the optimal design, \(\bm \theta^*\), via finding the optimal $\bm z^*$, which minimizes the robust compliance of the structure, \(Q_\text{rob}\), given the uncertain load angle. 

\subsubsection{Evaluation of the VAE reconstruction and generation}
\label{vae_eval}
In the first step, in order to build the search subspace using VAE, we need to  obtain the training data, which are deterministic optimal designs at various realizations of load angle calculated using the SIMP approach. In particular, we  considered 1000 load angle samples between $0$ and $\pi$ and obtained the corresponding deterministic designs using the 88 line topology optimization code in Matlab \cite{andreassen2011efficient}). 

As a result, this data set includes 1000 topology samples, with each data point consisting of 100 \(\times\) 100 pixels. Next, we need to train a VAE to encode these 10,000 dimensional topology data into a latent space with a much smaller dimensionality.  In this work, we removed the 10 top-ranked  topologies based on their compliances, i.e. the 10 geometries that yielded  the 10 lowest robust compliance values.  This was done in order to study whether the resulting search subspace  is capable of encoding topologies that are better performing than those seen in the training set.   Out of the rest 990 geometries, we used 90 images for testing and $n_{train}$ images for training, with $n_\text{train} =100, 200, \cdots, 900$ as will be explained later. Also, we  considered various latent space dimensions, specifically $|\bm z| = 2, 5,10$, in order to study how the dimensionality of the latent space would impact the performance of the representation and the subsequent optimization process. Since we use only high-resolution images ($100\times100$ dimension) for training, we refer to this network as single-fidelity VAE (as opposed to multi-fidelity VAE, introduced in Section~\ref{sec:mf}).

The training images are the input to the encoder network, which consists of an input layer, 7 fully connected hidden layers and an output layer. The output layer is the latent space layer. The number of nodes in the seven hidden layers are 5000, 2500, 1000, 500, 100, 50 and 10 respectively. Input layer has 10,000 nodes, same as the number of features from the input image (100\(\times\) 100 pixels). The number of nodes in the latent space layer depends on the chosen latent space dimension, \(|\bm z|\). ReLU function is used as the activation function for all the hidden layers.

The decoder network consists of an input layer, eight fully connected hidden layers, an AVG (average) pooling layer and an output layer. The input layer has \(|\bm z|\) number of nodes. The eight hidden layers in the decoder network has number of nodes 10, 50, 100, 500, 1000, 2500, 5000 and 10000 respectively. ReLU function is used as the activation function for all the hidden layers. The output of the eighth hidden layer, which has the same dimension as the input image, is passed through an AVG pooling layer with \(9\times9\) window.  The AVG pooling layer ensures that the output image does not have checkerboard pattern. 

\begin{figure} 
\centering
\includegraphics[width=0.6\linewidth]{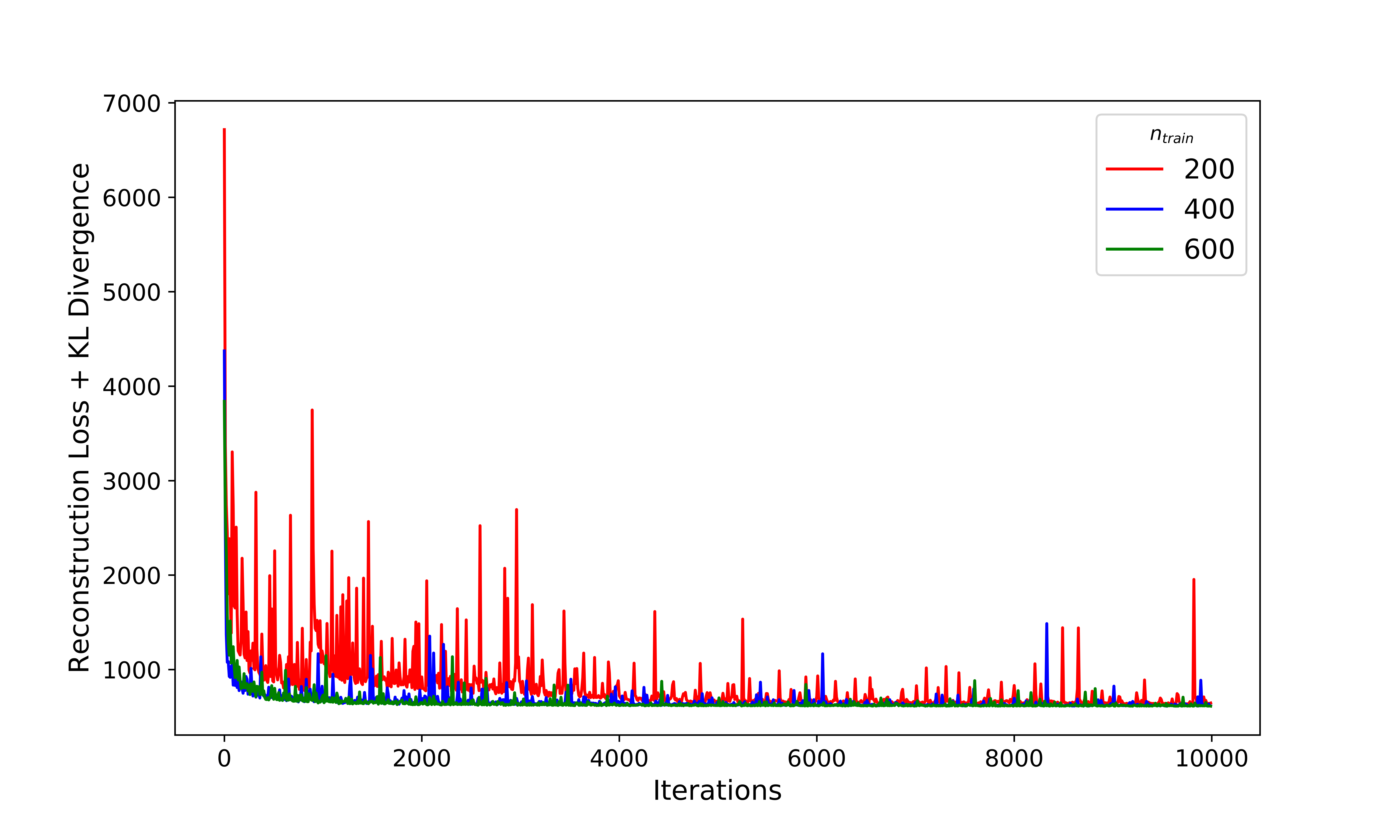}
\caption{Total testing loss for various training sample sizes, \(n_\text{train}\). Total loss is the sum of reconstruction error and KL Divergence. The results are shown for $|\bm z|=2$.}
\label{fig:loss_ss}
\end{figure}

\begin{figure} 
\centering
\includegraphics[width=0.6\linewidth]{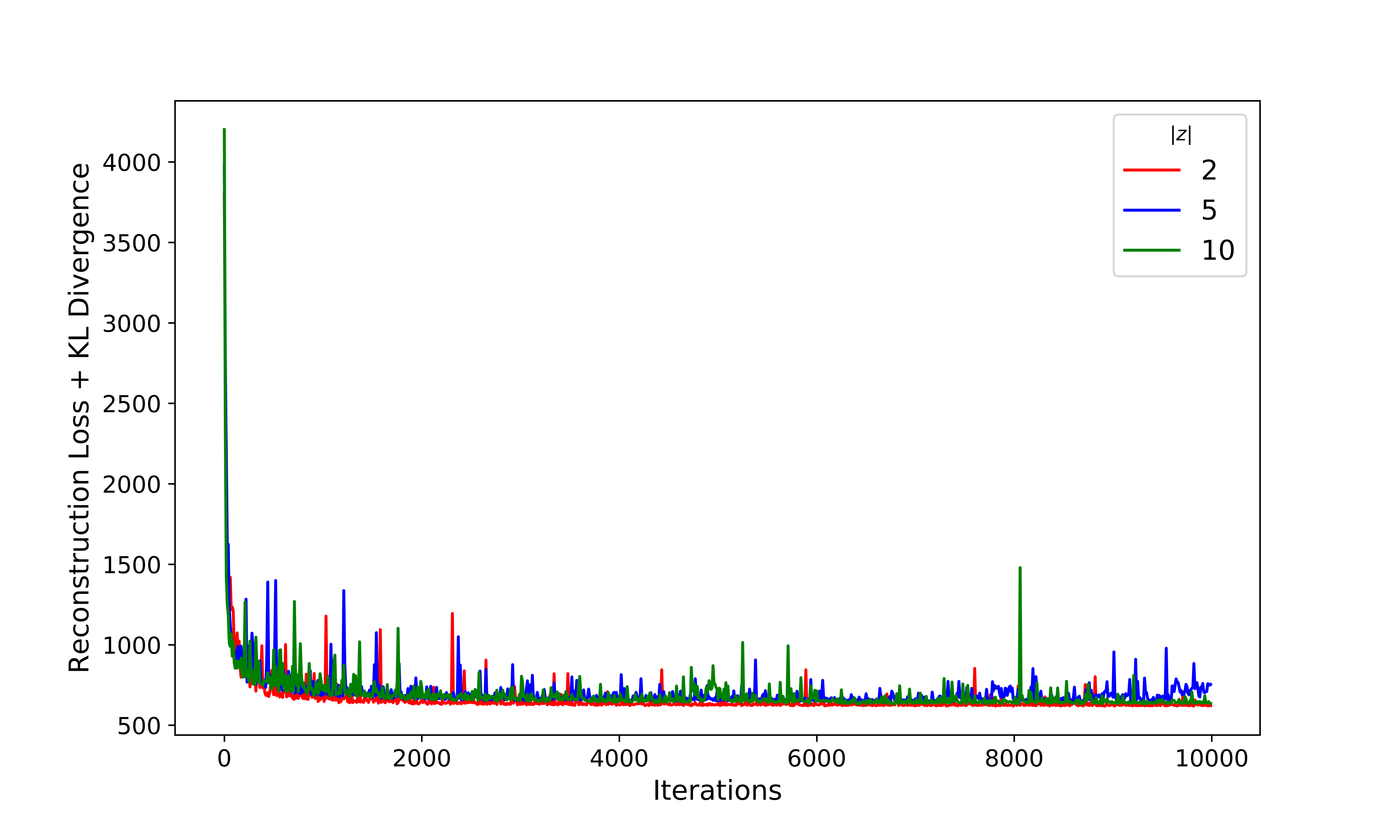}
\caption{Total testing loss for various latent space dimensions, \(|\bm z|\). Total loss used is the sum of reconstruction error and KL Divergence. The results are shown for $n_\text{train}=600$.}
\label{fig:loss_z}
\end{figure}

Fig.~\ref{fig:loss_ss} shows how different choice of \(n_\text{train}\) impact  the testing loss of single-fidelity VAE, for \(|\bm z|=2\). We can observe that with $n_\text{train}=400, 600$, the error converges more quickly compared to the case with $n_\text{train}=200$. However, their loss values do not vary much. In order to study the impact of latent space dimension, \(|\bm z|\), on the performance of VAE representation, we trained the model for  \(|\bm z| = 2, 5, 10\). Fig.~\ref{fig:loss_z} shows how the testing loss of single-fidelity VAE for different values of \(|\bm z|\), when training is done with \(n_\text{train}=600\). It can be observed that $|\bm z|$ does not have a noticeable  impact on the convergence of the training.

\begin{figure} 
\centering
\subfigure[Images from the testing dataset]{\includegraphics[width=0.8\textwidth]{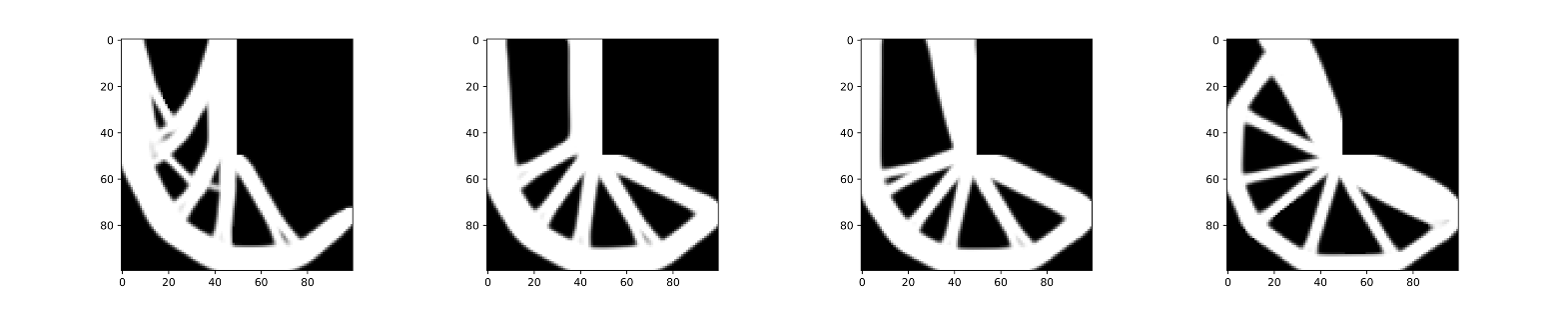}}\quad\\
\subfigure[Reconstructed images from single-fidelity VAE]{\includegraphics[width=0.8\textwidth]{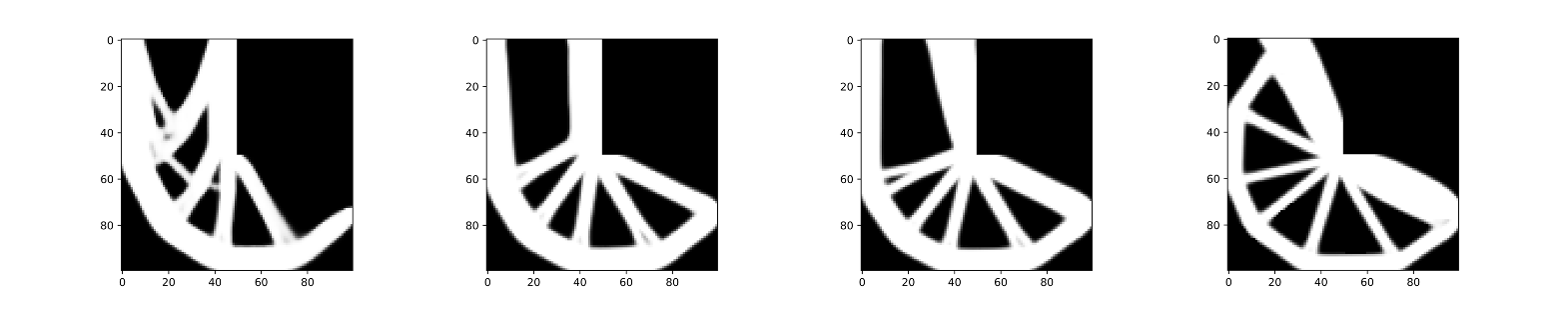}}\quad
\caption{Comparison between the test images (top) with their corresponding reconstructed images from single-fidelity VAE (bottom).}
\label{fig:test_reco} 
\end{figure}

\begin{figure} 
\centering
\includegraphics[width=0.6\textwidth]{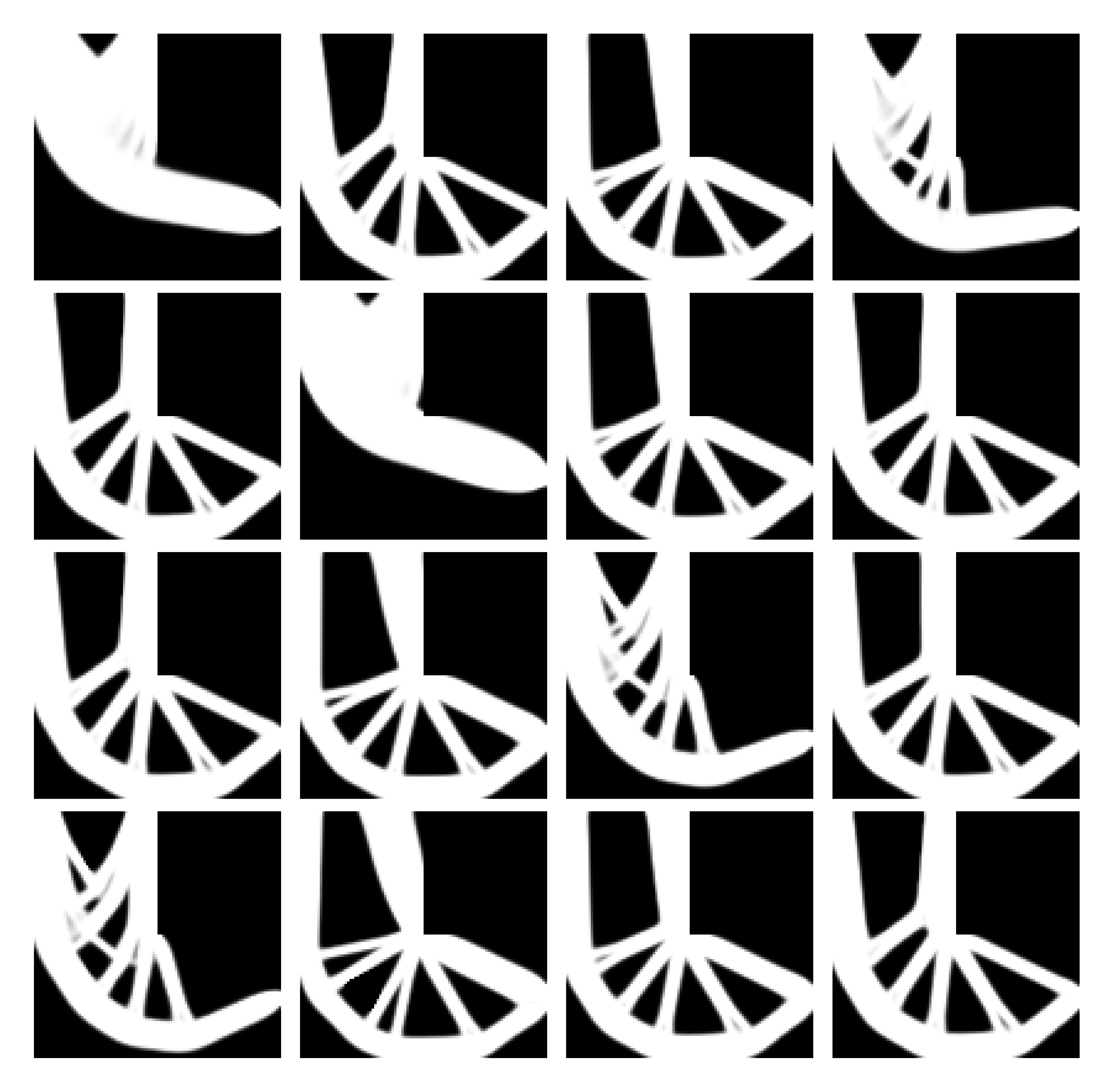}
\caption{A representative collection  of topologies generated by the decoder network of the VAE, via sampling from the standard normal $\bm z$ with $|\bm z|=2$   and $n_\text{train}=600$ using single-fidelity VAE. }
\label{fig:fake_images_ss} 
\end{figure}

As samples for the results, Fig.~\ref{fig:test_reco} shows four test images and their corresponding reconstructed images from VAE. As evident from the figure, the single-fidelity VAE reconstructs the original image with high accuracy. We can also generate new images, not present in the training or testing data set, by sampling from the prior distribution of \(\bm z\), and feeding it to the decoder network. Fig.~\ref{fig:fake_images_ss} shows a sample of 16 images generated by sampling \(\bm z\) from \(N(0,\bm I)\) for VAE with $n_\text{train}=600$. We can observe that the generated images are of good quality, closely resembling the overall structure of the training images, but with variation in the detailing of the design. A notable observation is that, even without imposing the volume fraction constraint of the optimization problem through the loss function, the network is able to learn this feature through training and is able to impose it on its output. In our experiments, only less than 1\% of the generated images from single-fidelity VAE has volume fraction, $\alpha_\text{v}>0.4$, with maximum volume fraction less than 0.405. Moreover, while the requirement of non-zero density at load points is also not explicitly fed into the model, we observe that model implicitly learns this property as observed from the generated images. Thus, while the design constraints of the problem are not explicitly checked during the training, the model is still able to successfully impose it by implicitly learning from the training data.  

\subsubsection{Evaluation of the compliance neural network surrogate}

\begin{figure} 
\centering
{\includegraphics[width=8cm]{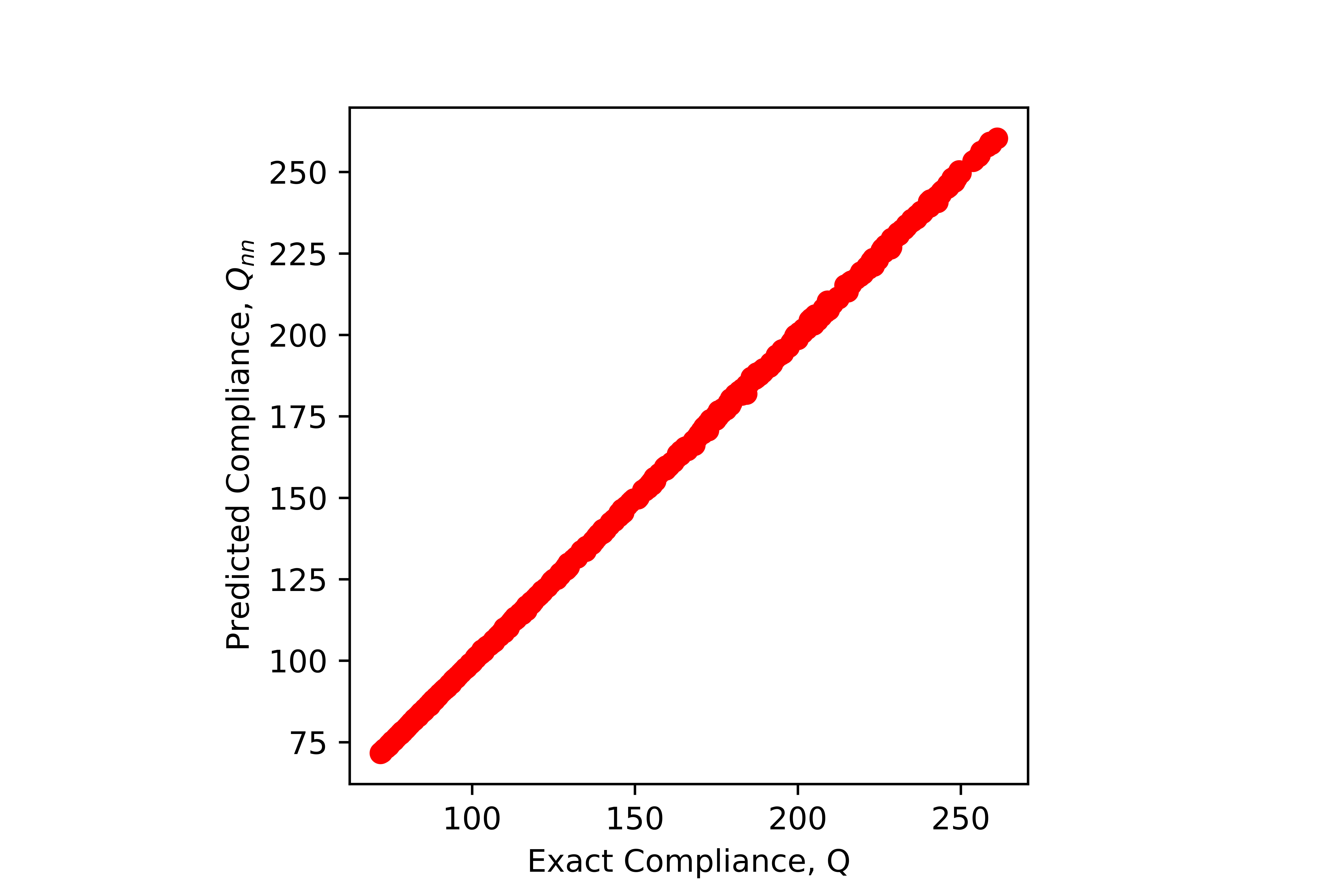}}
\caption{The comparison between the ``exact" values of robust compliance, $Q_{rob}$,  with the predicted robust compliance \(Q_\text{NN}\).  The training converged after 100 iterations where the loss function is the mean squared error between the ``exact" robust compliance calculated using quadrature method  and the predicted one.}
\label{fig:comploss}
\end{figure}

The compliance neural network surrogate, $h(\cdot)$, is a fully connected deep neural network with one input layer, seven hidden layers and an output layer. Input layer has 10,000 nodes (equal to the dimension of the topology images $\bm \theta$). The hidden layers have 5000, 2500, 1000, 500, 100, 50 and 10 nodes, respectively. Output layer is a single node which predicts the robust compliance value, \(Q_\text{NN}\).  Adam optimizer is used to minimize this loss function, with a constant learning rate of 1e-4. The compliance neural network is trained using the 1000 images generated using SIMP.  Both VAE and compliance neural network surrogates are trained simultaneously. The robust compliance values of the topologies are calculated using the  quadrature method~\cite{du1994application}. A total of seven quadrature points, sampled between 0 and 1 and scaled to be in the range of the load angle, and corresponding quadrature weights are used for calculating the robust compliance. For each topology, the deterministic compliance values, $Q$, are calculated for load angles corresponding to each of the quadrature points and they are averaged using the corresponding quadrature weights to calculate the robust compliance, $Q_\text{rob}$. Fig.~\ref{fig:comploss} shows the accurate prediction of the compliance surrogate after only  100 training iterations. 

\subsubsection{Evaluation of the gradient descent optimization}
\begin{figure} 
\centering
\subfigure[Contour plot of compliance distribution]{\includegraphics[width=7cm]{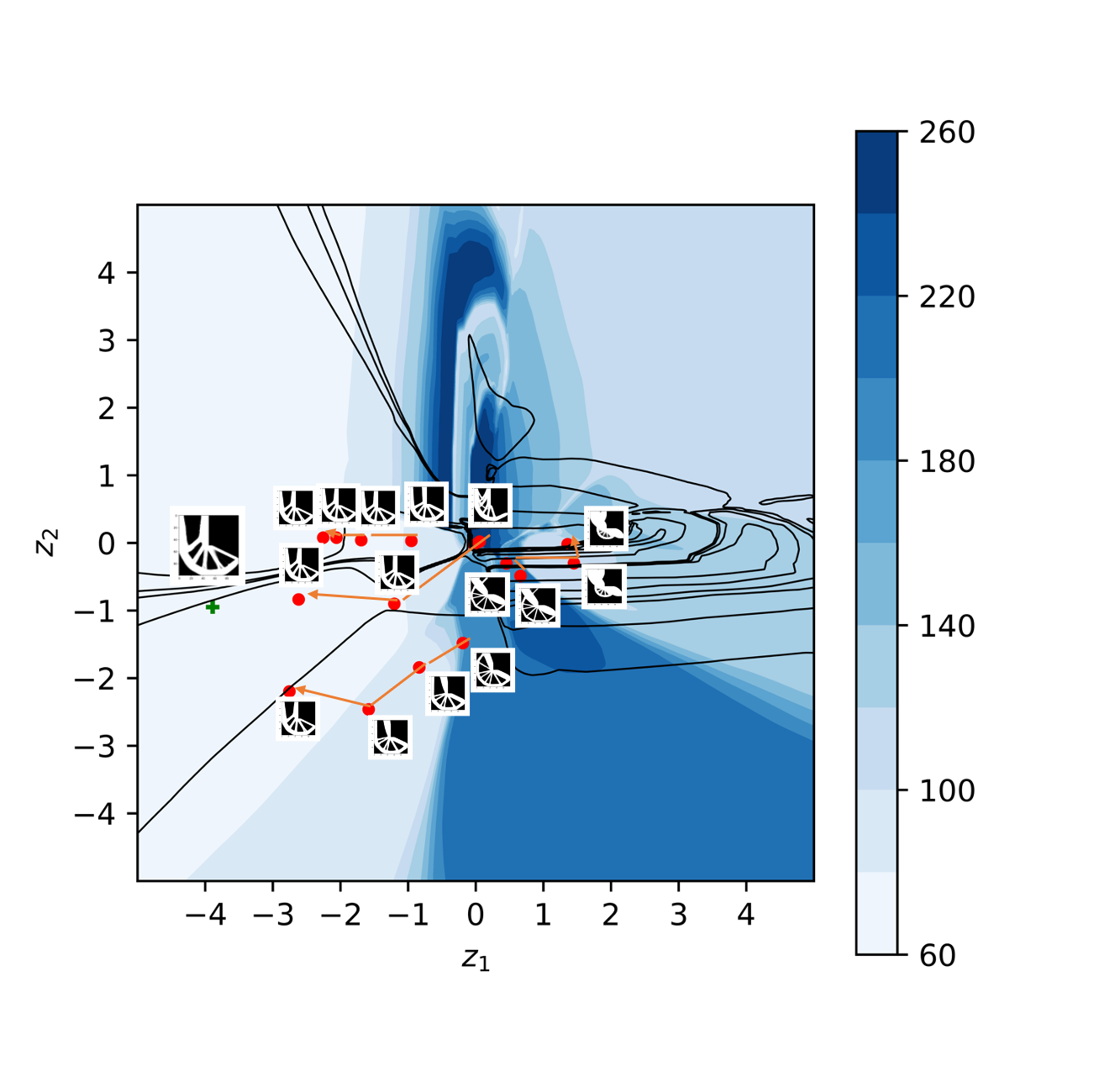}}\quad
\subfigure[Gradient descent evolution to the optimal design]{\includegraphics[width=9cm]{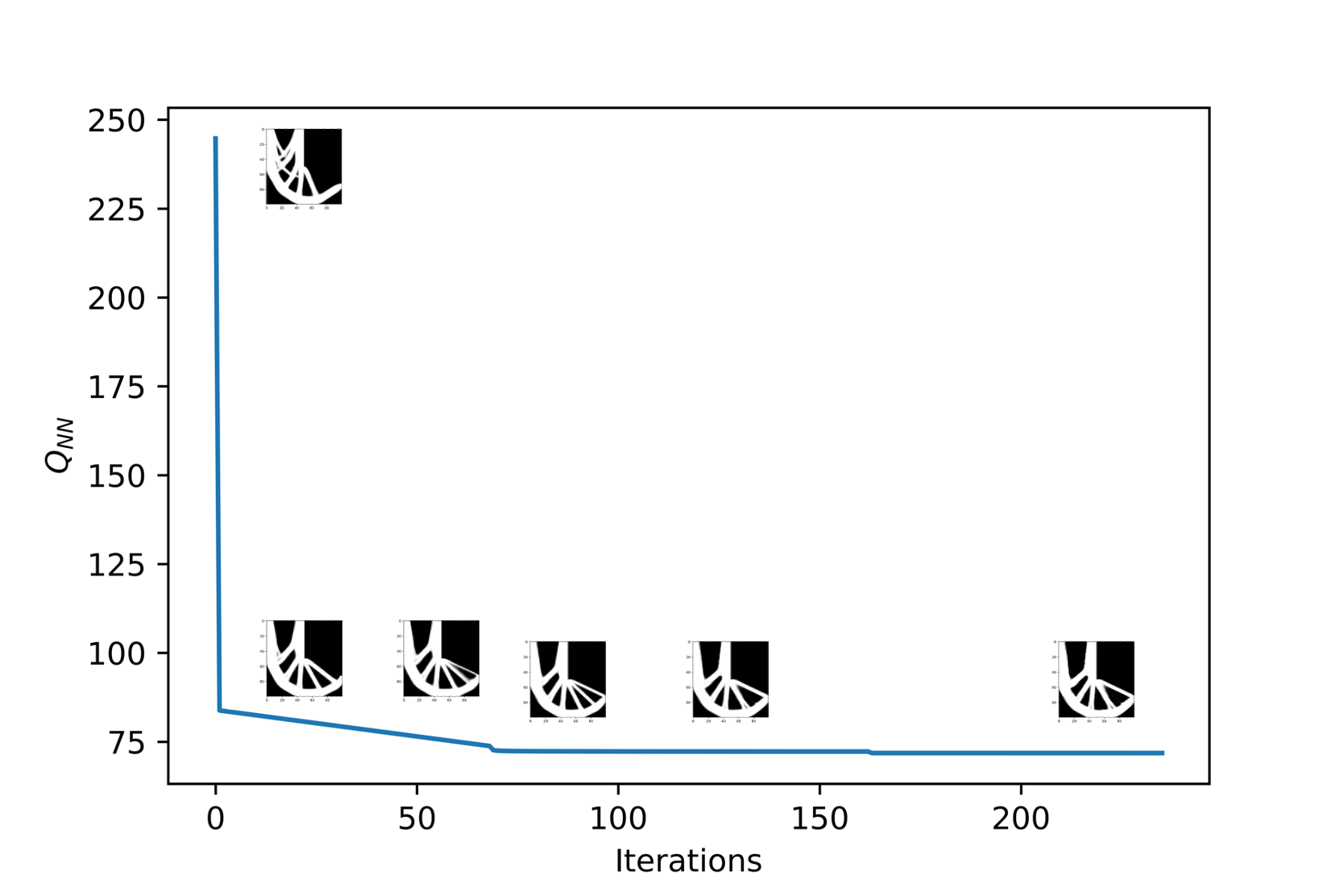}}\quad
\caption{Figure (a) shows a contour plot of robust compliance values in the latent space for $|\bm z|=2$ along with the gradient descent evolution of the designs from multiple starting points. The  + sign indicates the location of the (global) optimal design. It can be observed that gradient descent could lead to local minima for certain starting points. Starting with a good $\bm z$ value can avoid falling into local minima. Figure (b) shows the gradient descent evolution from the starting image to the best design. }
\label{fig:journey}
\end{figure}

Using the trained single-fidelity VAE and compliance surrogates, we solve the RTO problem by solving the unconstrained problem in Eq.~(\ref{eq.zRTO}). This involves   applying gradient descent algorithm in the latent space (as shown in Fig.~\ref{fig:nnarch}). One of the shortfalls of gradient descent is the possibility of designs to optimize to local minima instead of global minima. Fig.~\ref{fig:journey} shows how starting at different $\bm z$ values can lead to different local minima. In order to avoid this and to start with a good initial design  \(\bm z_0\) , we use a brute-force approach where we draw  a moderate number of \(\bm z\) samples and evaluate their approximate robust compliance, \(Q_\text{NN}\), and select the design with the lowest compliance as the initial  point for the gradient descent step. 

\begin{figure} 
\centering
\subfigure[Deterministic optimum, \(Q_\text{rob}=71.59\)]{\includegraphics[width=0.38\textwidth]{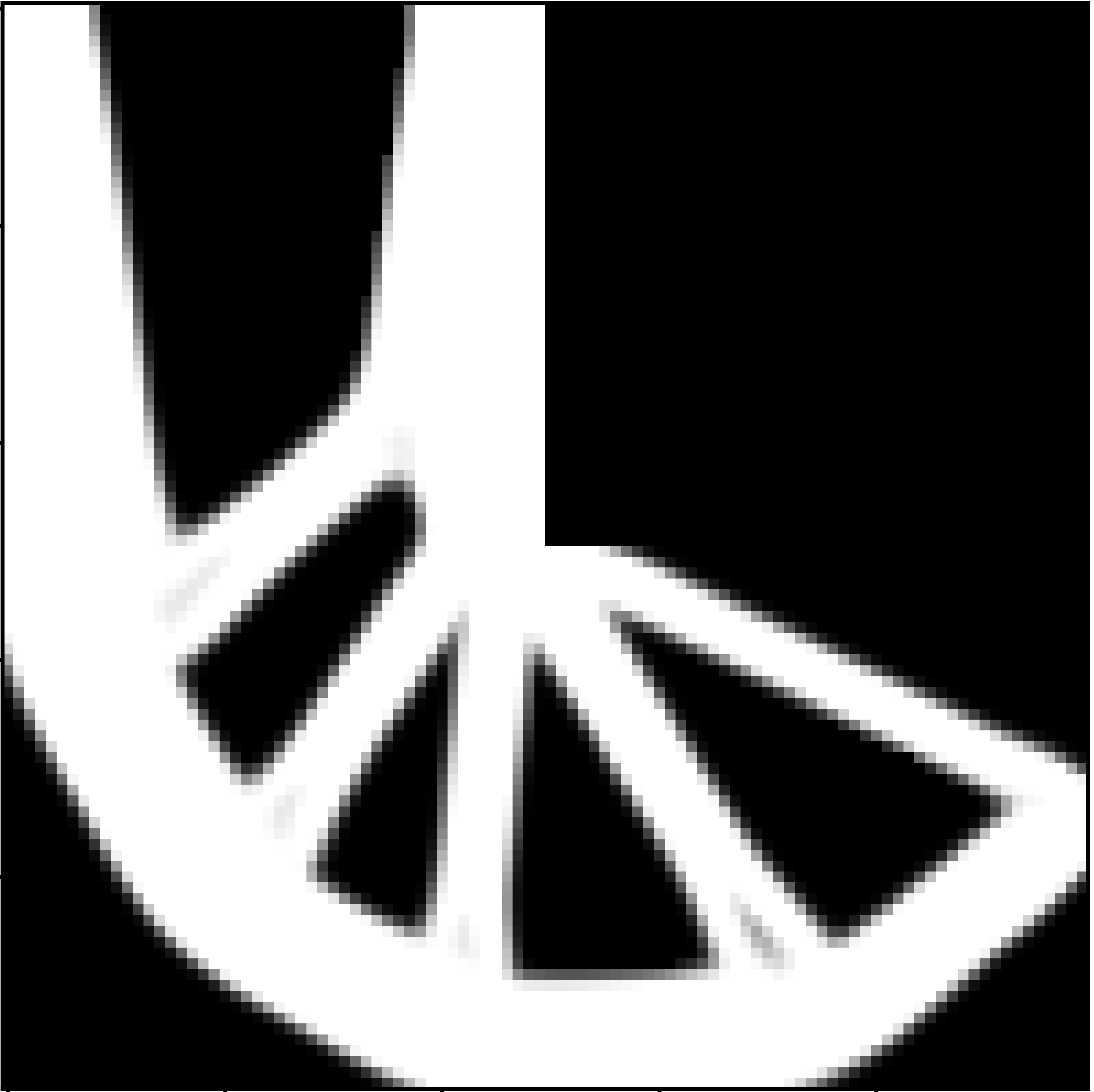}}\quad
\subfigure[Best training sample($n_\text{train}=200$), \(Q_\text{rob}=72.02\)]{\includegraphics[width=0.38\textwidth]{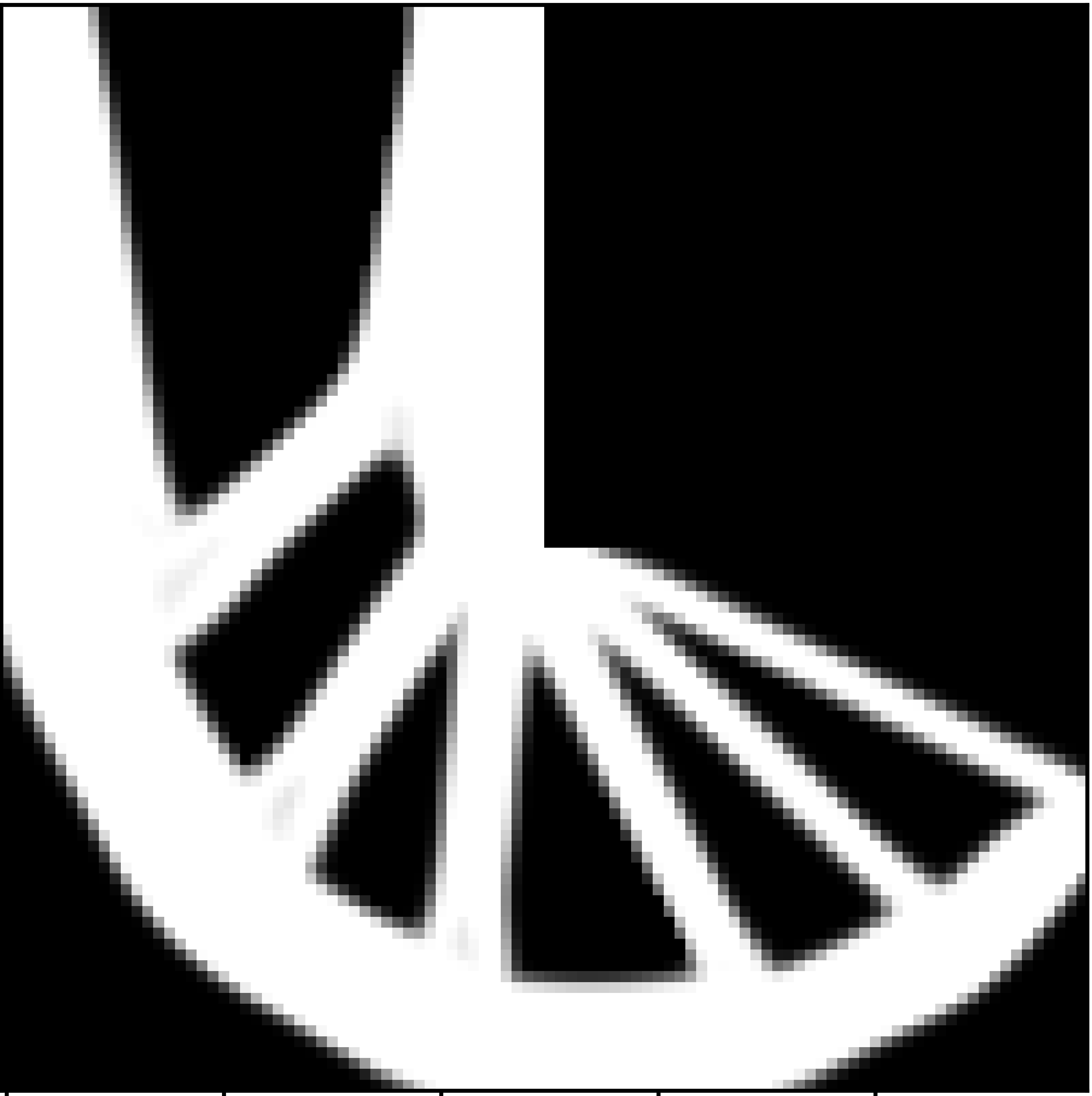}}\quad 
\subfigure[Best training sample ($n_\text{train}=600$), \(Q_\text{rob}=71.89\)]{\includegraphics[width=0.38\textwidth]{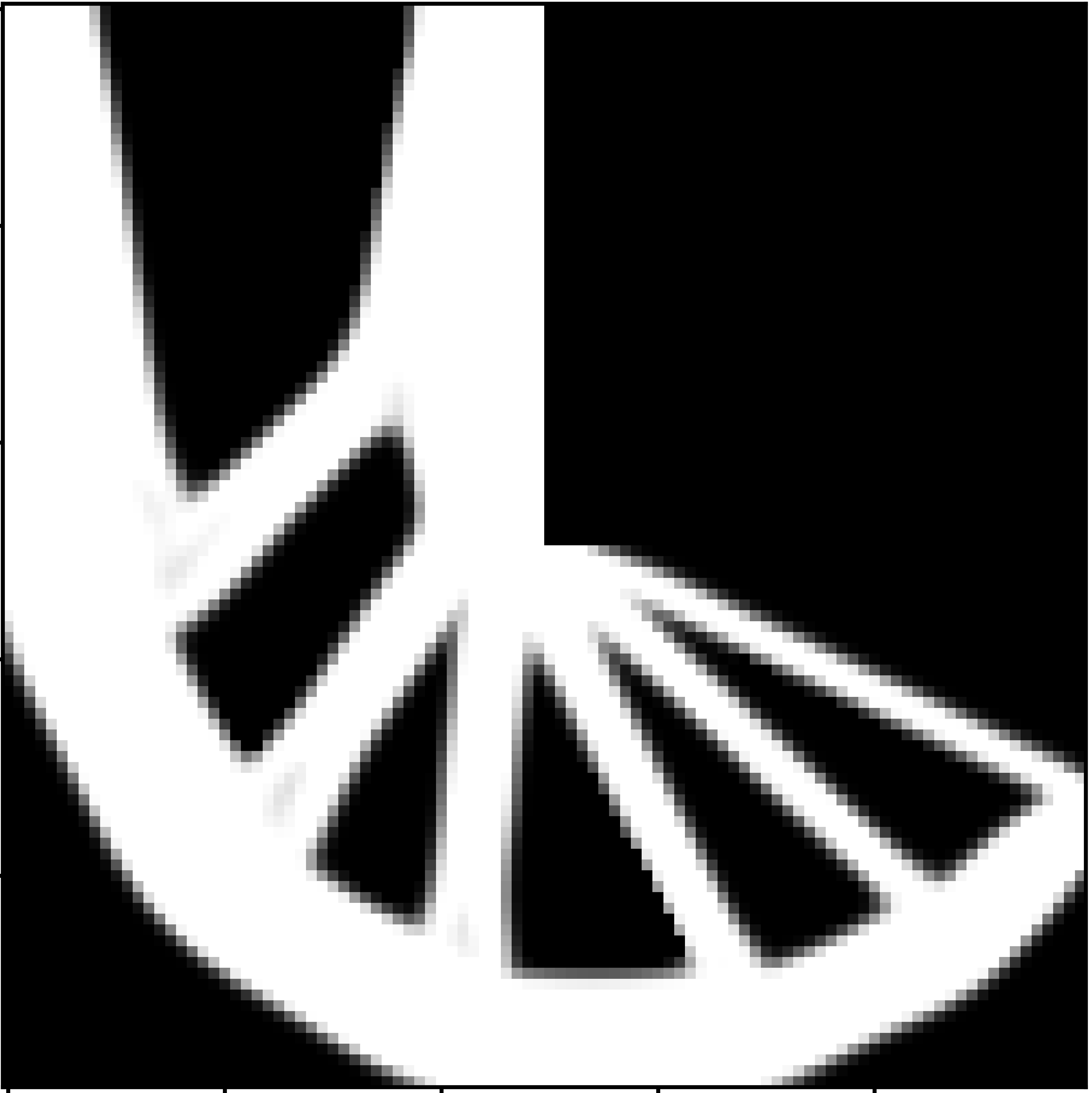}}\quad\\
\subfigure[Single-fidelity robust optimum ($n_\text{train}=200$), \(Q_\text{rob}=71.74\)]{\includegraphics[width=0.38\textwidth]{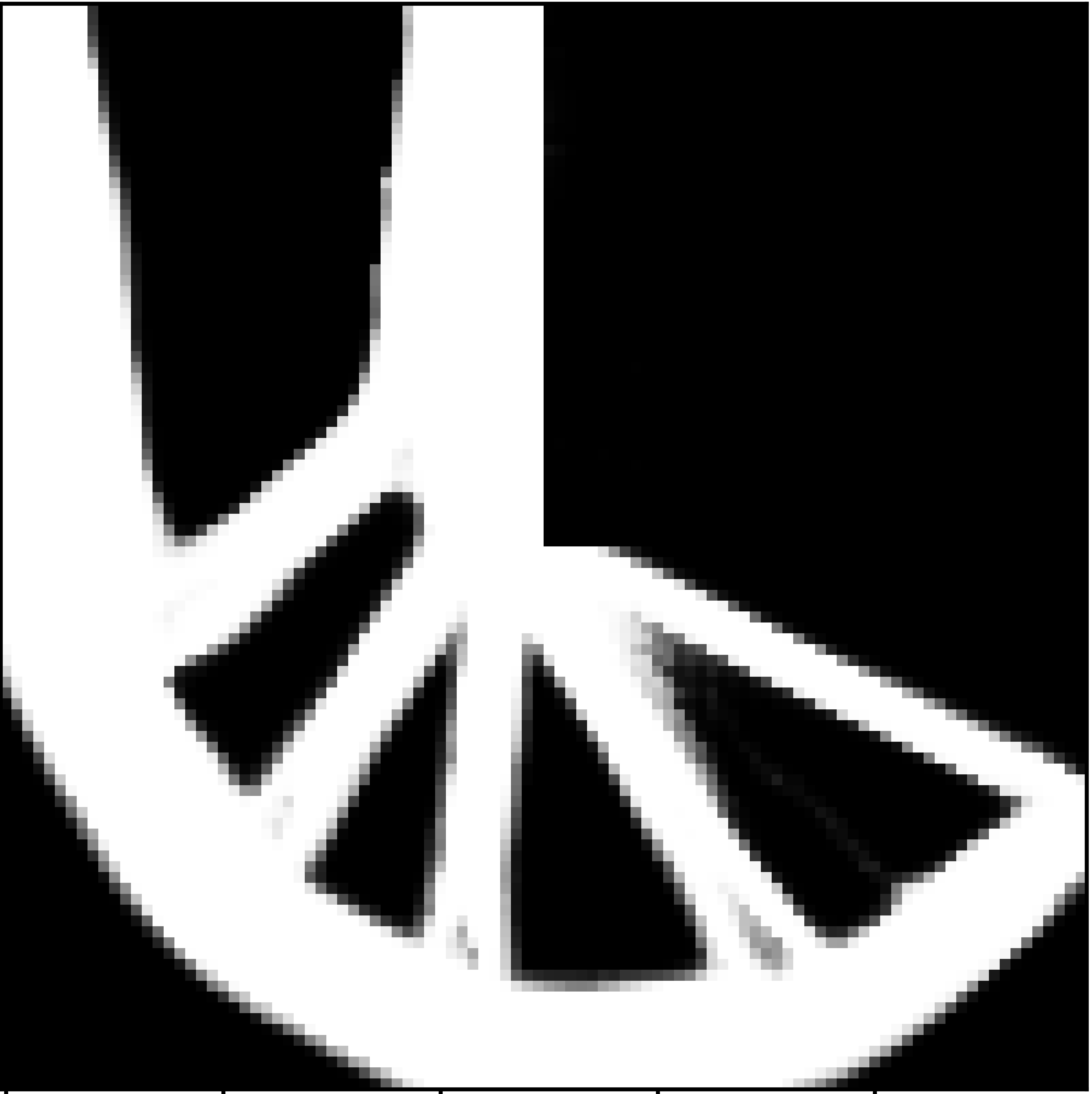}}\quad
\subfigure[Single-fidelity robust optimum ($n_\text{train}=600$), \(Q_\text{rob}=69.84\)]{\includegraphics[width=0.38\textwidth]{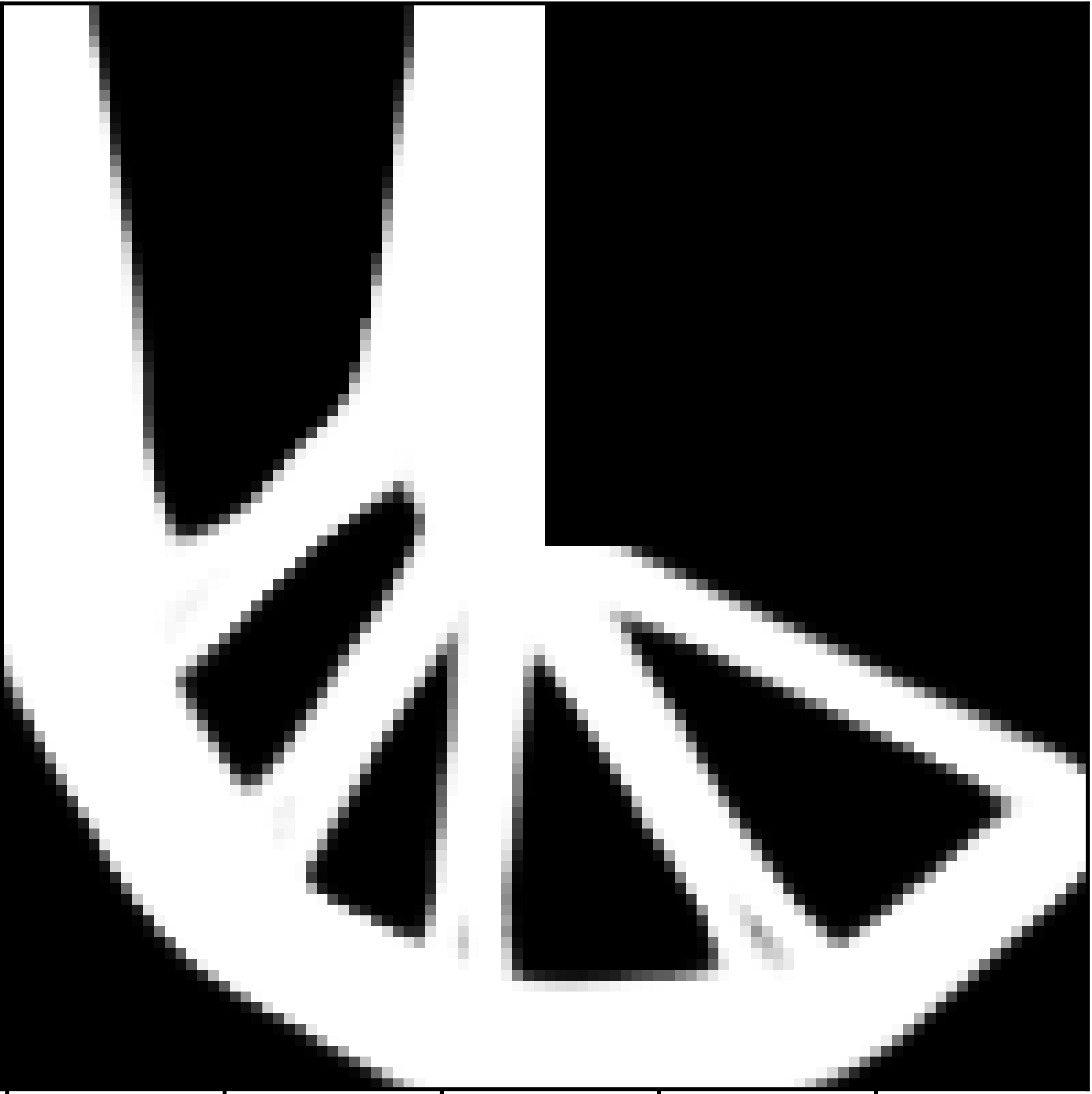}}\quad
\subfigure[Multi-fidelity robust optimum for low-resolution ($n_\text{train}=700$), \(Q_\text{rob}=78.64\)]{\includegraphics[width=0.38\textwidth]{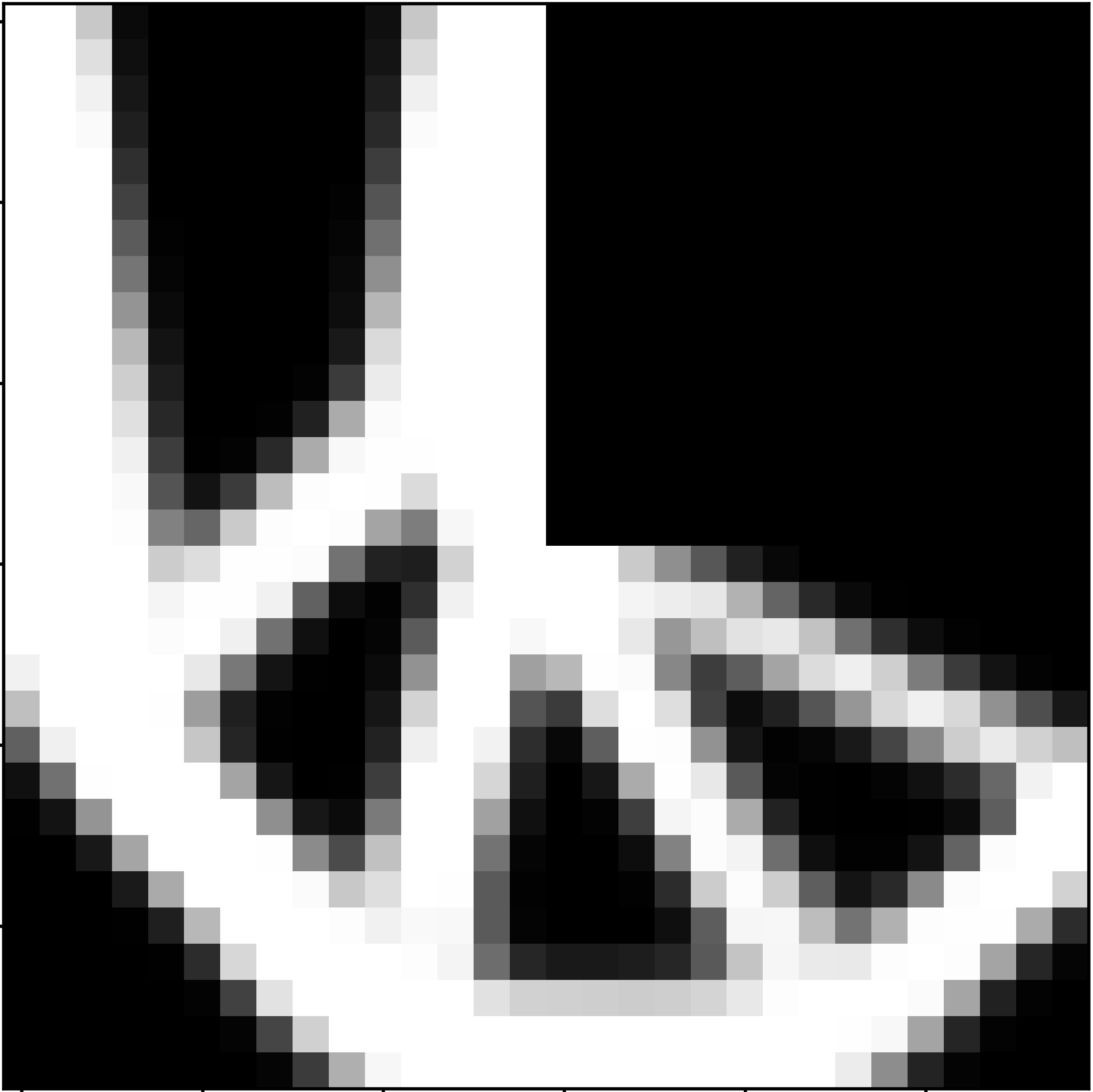}}\quad
\subfigure[Multi-fidelity robust optimum for high-resolution ($n_\text{train}=20$), \(Q_\text{rob}=69.09\)]{\includegraphics[width=0.38\textwidth]{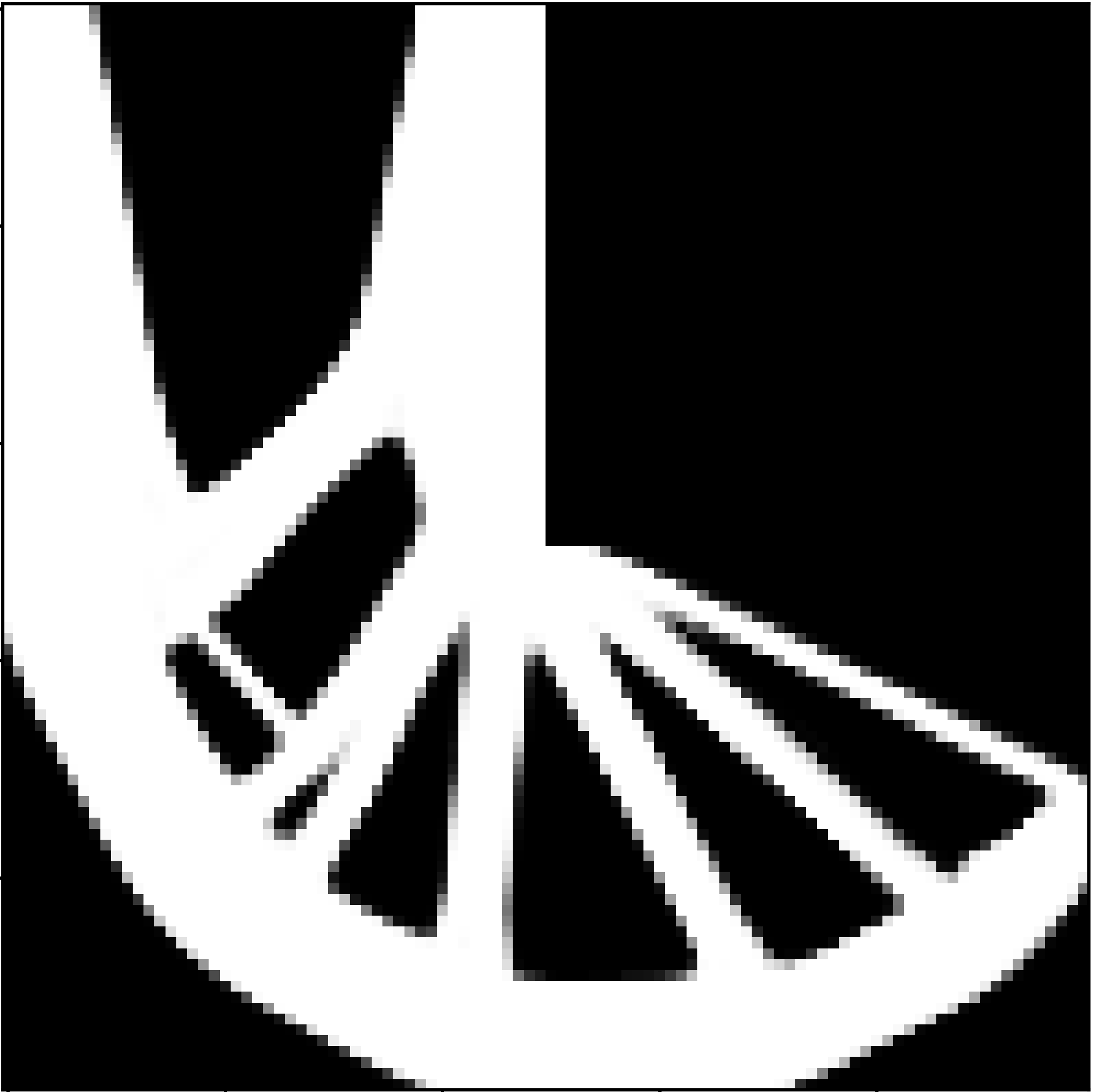}}\quad\\
\caption{Comparison of the most optimal designs through various sources}
\label{fig:optimaldesign}
\end{figure}

Fig.~\ref{fig:optimaldesign} shows the optimal designs obtained through this process where 100 initial samples were used to determine the best initial design. Out of the 1000 deterministic designs obtained from SIMP, the best one had a robust compliance of $Q_\text{rob}=71.59$. As mentioned earlier, the top ten deterministic designs (based on the robust compliance value) are not included in the training of the VAE. Here, we have included the results of two single-fidelity VAEs trained with $n_\text{train}=200$ and $n_\text{train}=600$. Both the models had latent space dimension of $|\bm z|=2$, which showed the same performance levels compared to higher dimensions. The single-fidelity VAE with $n_\text{train}=200$ is computationally much faster to train and the resulting gradient descent produces a design that is better than the training samples. The single-fidelity VAE with $n_\text{train}=600$ produced a better design via the gradient descent approach, with $Q_\text{rob}=69.84$ which was smaller than $Q_\text{rob}=71.89$ observed in the training data. This design is better, in terms of the robust compliance, than the best design from the deterministic optimal space as well.  Fig.~\ref{fig:journey} shows the evolution of the gradient descent from the starting image to its optimal design. These results underscore the effectiveness of the approach to efficiently explore the search subspace and its capability to improve the design beyond the training set. It should however be noted that it cannot still outperform the best robust design obtained from SIMP mainly because a smaller subspace is being explored.

\subsubsection{Evaluation of multi-fidelity approach on finding the optimal designs}
\label{sec:mf}

In the multi-fidelity approach, we use  700 low-resolution images for training both low-fidelity VAE and low-fidelity compliance NN surrogate. The low-resolution is chosen to be $30 \times 30$. High-fidelity networks use $100 \times 100$ images, which are the same images used in the single-fidelity networks of the previous sections. The data  generation process is the same as that for the single-fidelity VAE. For the training of neural networks, we remove the top 10 topologies with the highest compliances, similarly to the process prescribed in Section~\ref{vae_eval}. 

The low-fidelity VAE architecture consists of a encoder network with a 900-dimensional input layer (related to $30 \times 30$  images) and 4 hidden layers with 500, 100, 50 and 10 nodes, respectively. This is followed by a latent  space layer, with $|z| = 2$. The decoder network consists of 4 hidden layers with 10, 50, 100 and 500 nodes, leading to the 900-dimensional output layer. ReLU function is  used as the activation function for all the layers.

The high-fidelity VAE is built by adding two layers to the beginning and two layers to the end the low-fidelity VAE. Thus, the high-fidelity encoder would consist of an input layer with 10,000 nodes followed by 6 hidden layers with 5000, 900, 500, 100,  50 and 10 nodes, and its six decoder layers have 10, 50, 100, 500, 900 and 5000 nodes leading to the 10,000 dimensional output layer. This multi-fidelity VAE architecture will have the same number of hidden  layers as the single-fidelity VAE with high-resolution topologies that was used in Section 4.1.1. We train the multi-fidelity VAE using 20 high-resolution images. In the training, we initialize the model  parameters for the first and last two layers with Xavier initialization, and the remaining parameters (i.e. those belonging to the pre-trained low-fidelity part) are initialized with trained values obtained for the low-fidelity model. 

For the compliance NN surrogates, we follow similar data generation and training processes. The low-fidelity compliance network has an input layer with 900 nodes and four hidden layers with 500,  100, 50 and 10 nodes and an output layer with a single node. Then, the high-fidelity architecture is created by adding two layers to the beginning of the low-fidelity  network, forming one input layer with 10,000 nodes, and six hidden layers  with 5000, 900, 500, 100, 50 and 10 nodes, with the same output layer. Similarly to multi-fidelity  VAE, we use 700 samples for low-fidelity training and 20 samples for high-fidelity training. All the  networks (i.e. VAE and compliance networks) are trained for 5000 iterations using Adam optimizer and a learning rate of 1E−4.

To discuss the multi-fidelity results, let us use the following notations for clarity. Single-fidelity VAE, which is the VAE trained with 600 high-resolution ($100\times100$) samples, is referred to as VAE-HF. The high-fidelity VAE trained using multi-fidelity approach with 20 samples of $100\times100$ resolution and pre-trained parameters from low-fidelity network is referred to as VAE-MF. Single-fidelity VAEs trained with different number of high-resolution samples, say $k$, can be differentiated by using the notation, VAE-HF-k. For example, single-fidelity VAE trained using 20 and 600 high-resolution samples can be differentiated as VAE-HF-20 and VAE-HF-600 respectively. 

\begin{figure} 
\centering
\subfigure[VAE-HF-20]{\includegraphics[width=0.45\textwidth]{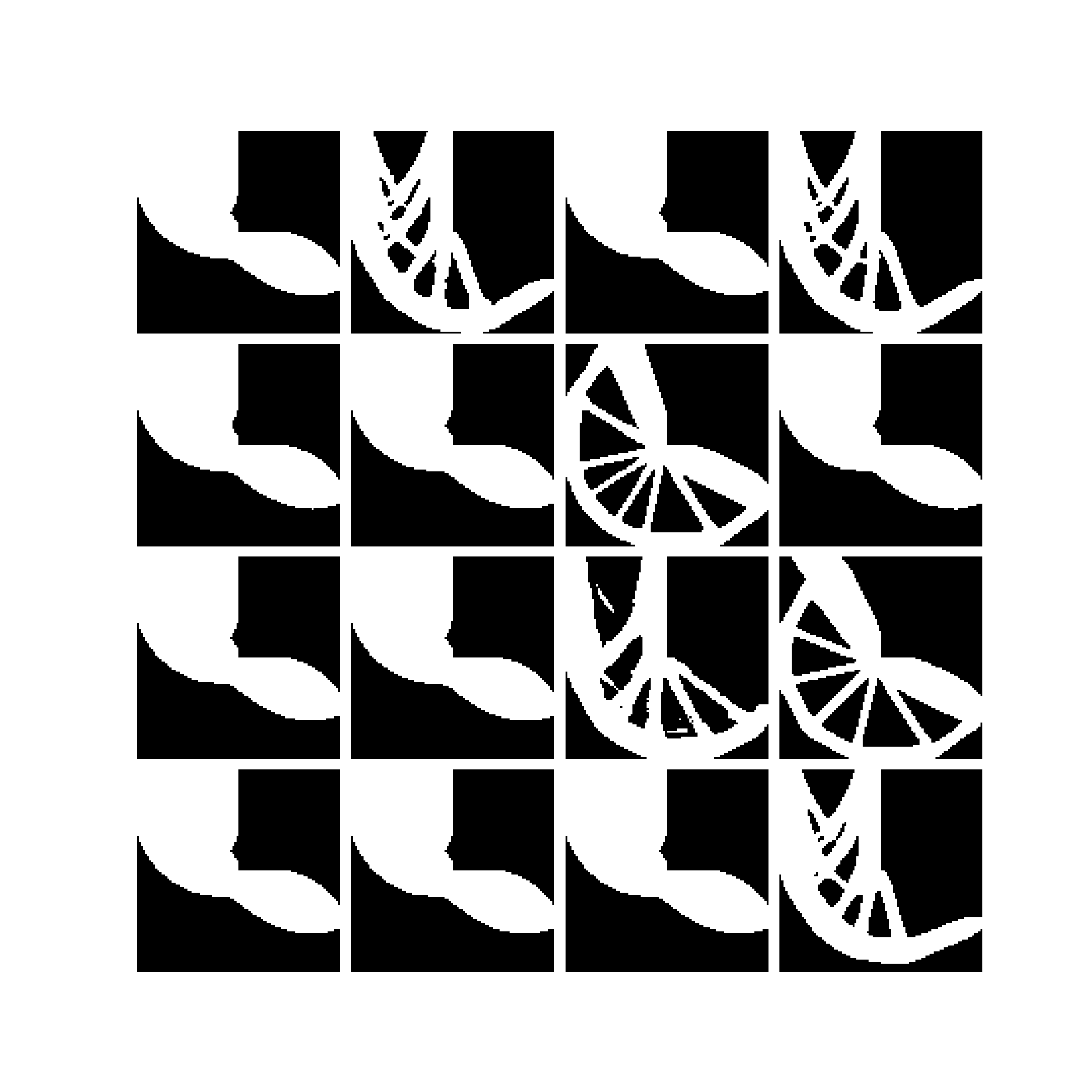}}\quad
\subfigure[VAE-MF]{\includegraphics[width=0.45\textwidth]{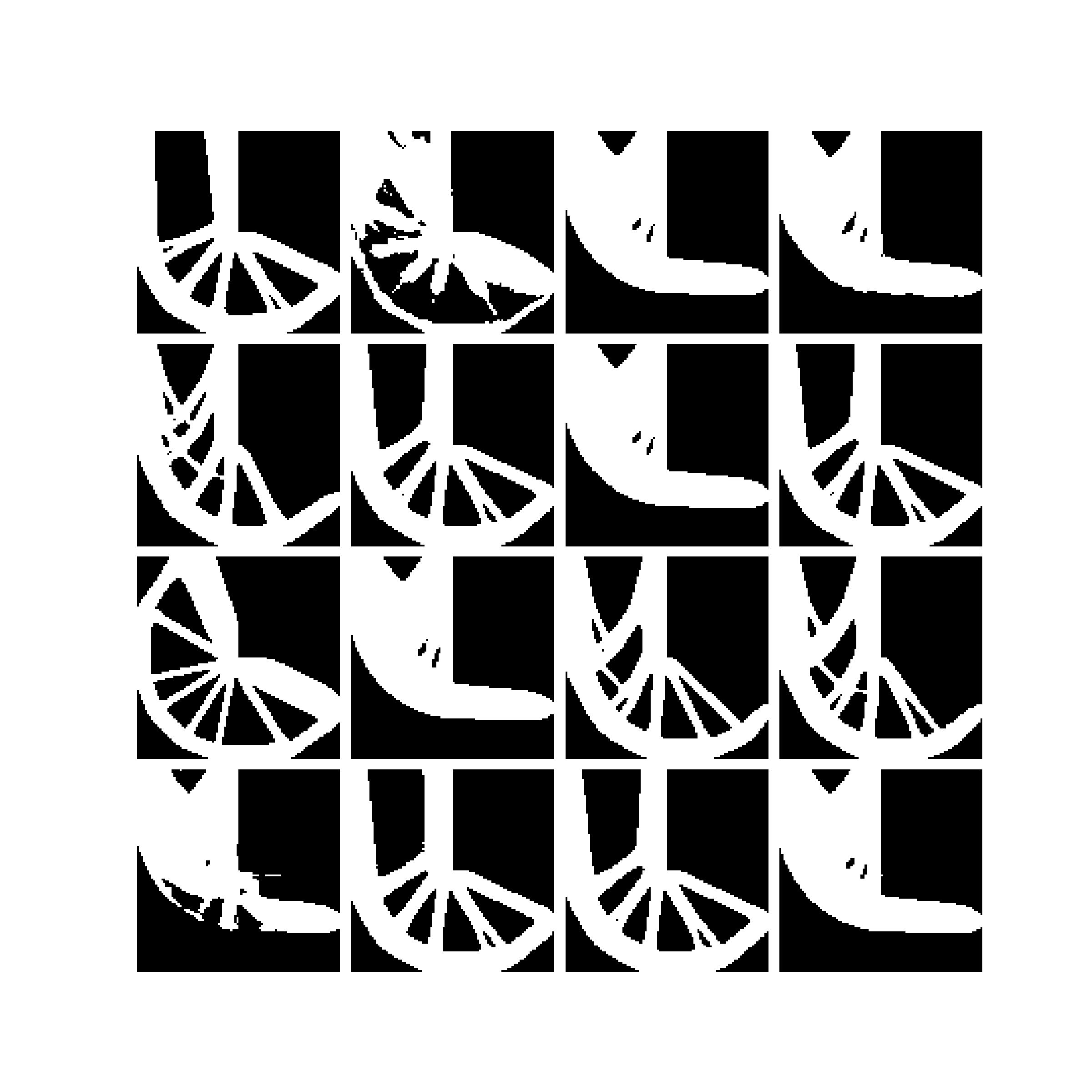}}\quad 
\caption{A sample of geometries generated by single- and multi-fidelity VAEs by randomly sampling from standard normal distribution and feeding it to the decoder network. Both the VAEs are trained using 20 high-resolution samples.}
\label{fig:random_samples_mf}
\end{figure} 

Fig.~\ref{fig:random_samples_mf}(b) shows a set of topologies generated by random sampling from the latent space using VAE-MF trained using 20 high-resolution images. On the other hand, the images shown in Fig.~\ref{fig:random_samples_mf}(a) are generated by VAE-HF-20, trained with the same 20 training samples, but with no trained parameters from low-fidelity VAE used for model initialization. It can be observed that the variability in the images generated by VAE-MF are much higher than that of VAE-HF-20. This indicates that the multi-fidelity model is able to generate additional variability beyond the one existing in the single fidelity VAE trained using 20 samples. 

\begin{figure} 
\centering
\subfigure[Reconstruction loss]{\includegraphics[width=0.45\textwidth]{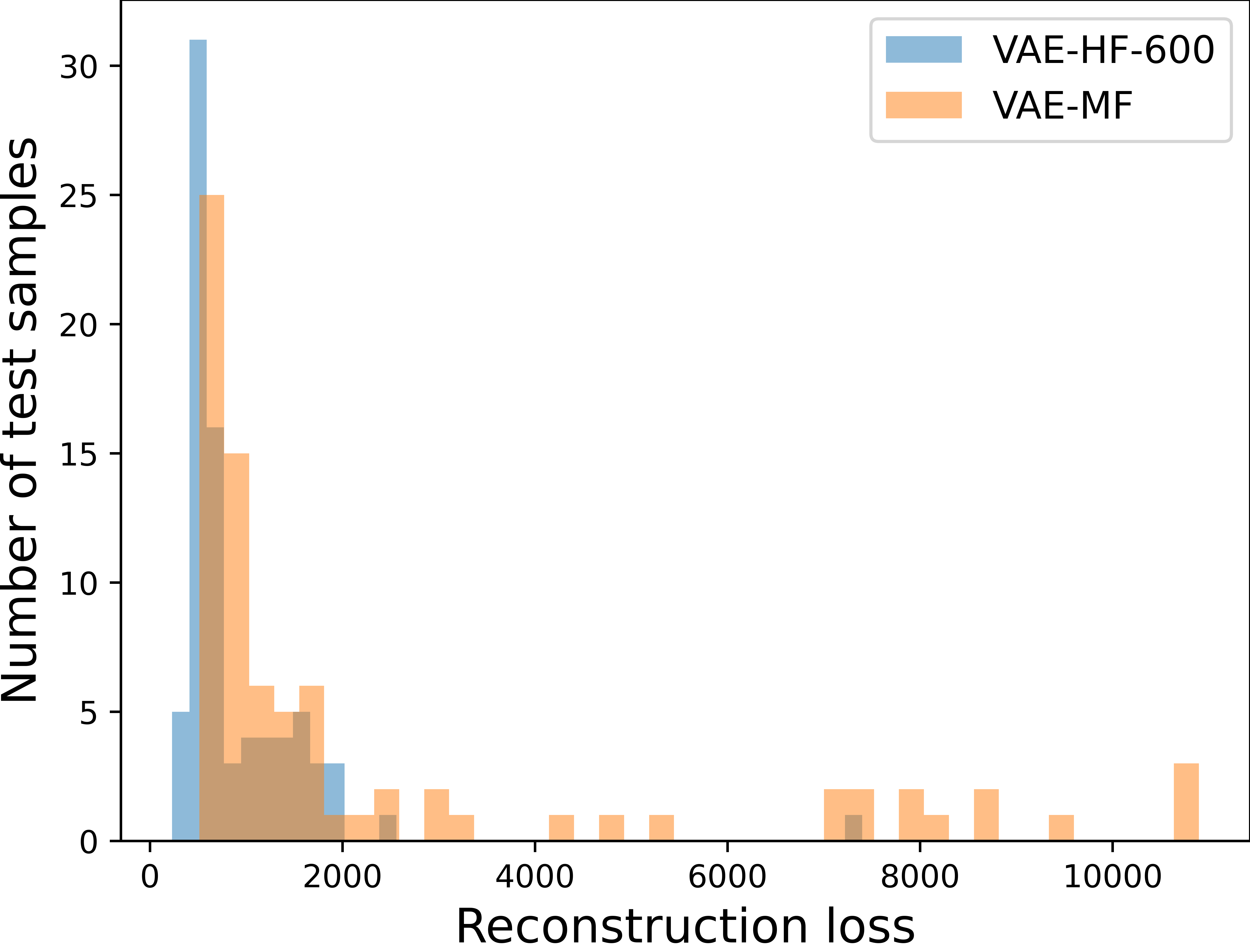}}\quad
\subfigure[Compliance loss]{\includegraphics[width=0.45\textwidth]{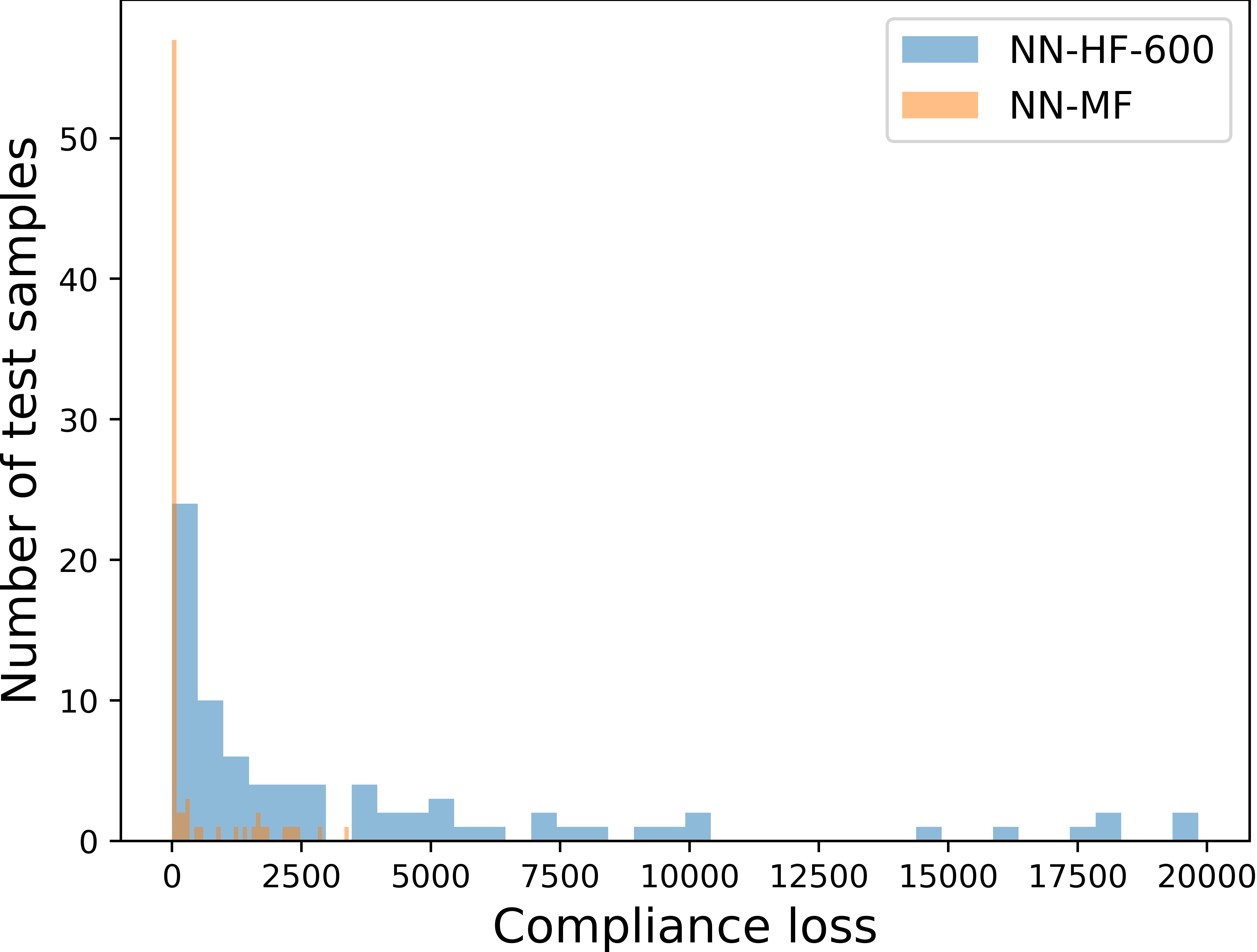}}\quad 
\caption{Histograms showing the distribution of reconstruction and compliance losses for the testing samples for single- and multi-fidelity  VAE and compliance NNs. VAE-HF-600 refers to single-fidelity VAE, trained using 600 samples of high-resolution $100\times100$ in Section \ref{vae_eval}. Similarly NN-HF-600 is the single-fidelity compliance NN trained in Section \ref{vae_eval}. VAE-MF and NN-MF refer to the high-fidelity networks in the multi-fidelity approach trained using 20 samples of resolution $100\times100$. All the networks are tested on 80 samples of $100\times100$ resolution.}
\label{fig:hist_mf}
\end{figure} 

\begin{figure} 
\centering
{\includegraphics[width=0.7\textwidth]{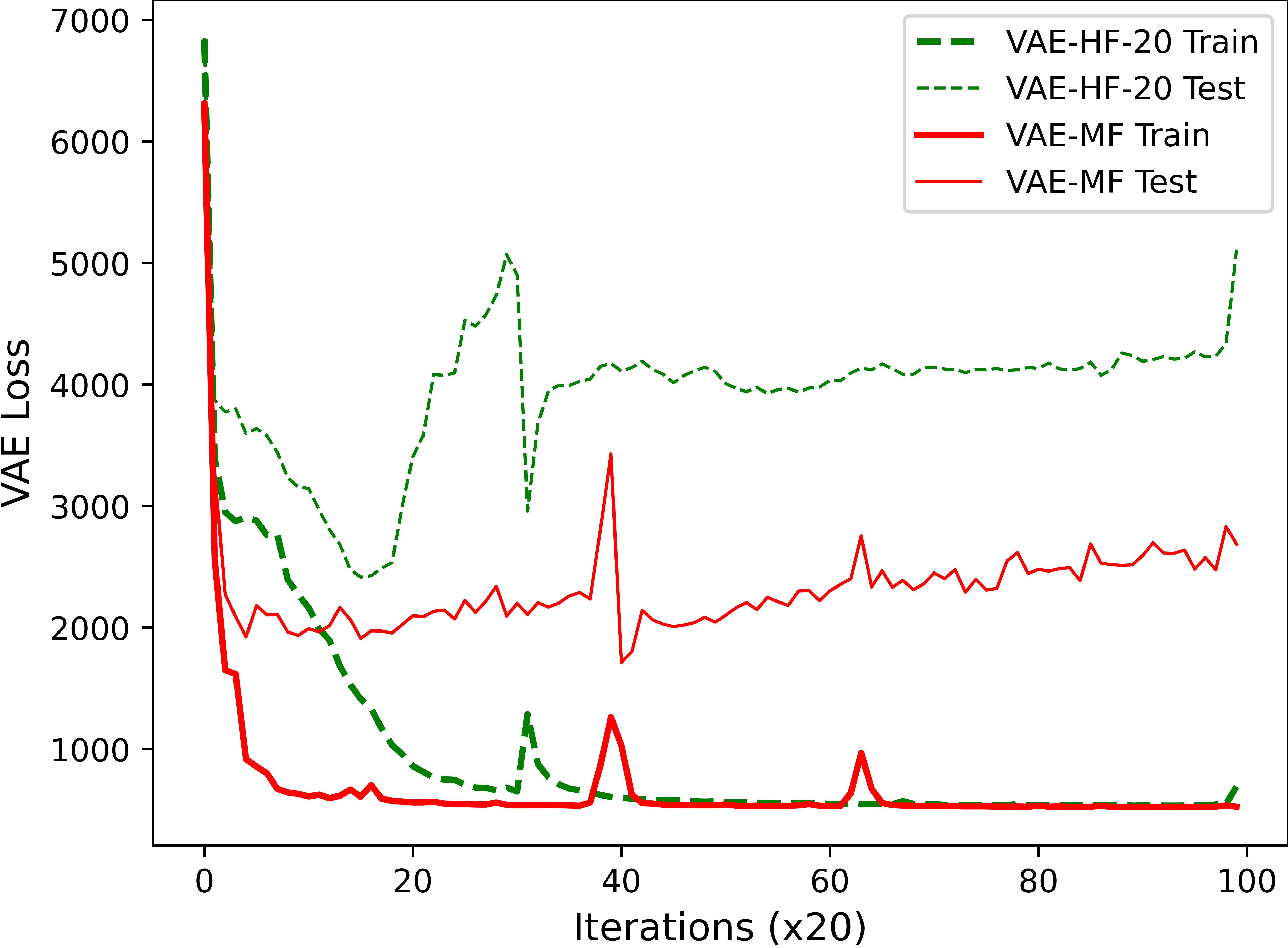}}
\caption{A convergence plot showing the total VAE loss on the training and testing data for VAE-HF-20 and VAE-MF. The training for both the networks are done using 20 images of $100\times100$ resolution and tested on 80 images of the same resolution.}
\label{fig:vaeloss_mf_sf20}
\end{figure}

The VAE loss for VAE-HF-20 and VAE-MF are also compared to learn the impact of using pre-trained parameters on the model convergence and generalization as shown in Fig.~\ref{fig:vaeloss_mf_sf20}. We can observe that VAE-MF converges very quickly, around 30 iterations, whereas, VAE-HF-20 converges around 800 iterations. Moreover, VAE-HF-20 has higher testing errors compared to that of VAE-MF, pointing to the better generalization capability of VAE-MF in reconstructing images not seen during the training. We also compare the reconstruction power of VAE-MF with that of the single-fidelity VAE, VAE-HF-600, trained using 600 high-resolution images. The histogram showing the distribution of the reconstruction errors for the 80 test samples for both the models are shown in Fig.~\ref{fig:hist_mf}(a). While VAE-HF-600 has better reconstruction errors compared to VAE-MF, the errors are still comparable and more than 90\% of the samples have reconstruction errors closer to that of VAE-HF-600. Fig.~\ref{fig:hist_mf}(b) shows the histogram of the distribution of the compliance prediction errors for single- and multi-fidelity compliance NN surrogates (referred to as NN-HF-600 and NN-MF respectively). We can observe that the multi-fidelity network has much better compliance prediction errors compared to that of the single-fidelity network.

Once the multi-fidelity VAE and compliance NN surrogates are trained, they are used for topology optimization using the gradient descent algorithm as detailed in Section~\ref{GD}. Fig.~\ref{fig:optimaldesign}(g) shows the optimal design obtained through the multi-fidelity approach. The best design among the 20 samples in the high-fidelity training data has a compliance value of 70.85. The optimal design identified via the multi-fidelity approach with a robust compliance value of 69.09 is better than those from the single-fidelity methods. We have also included the optimal design from the low-resolution network (trained with $30\times30$ samples) in Fig.~\ref{fig:optimaldesign}(f) to demonstrate the difference in the optimal designs between low- and high-resolution networks in multi-fidelity approach. This shows that the optimal design from the high-fidelity network is not the same design (but with a better resolution) as that from the low-fidelity model, but rather it is a new design generated by the high-fidelity VAE. The largest gain through this approach is observed in the computational efficiency in data generation and model training. The training data generation for 600 samples of resolution $100\times100$ takes 3.55 hours, whereas it takes only 34 minutes to generate 700 low resolution samples and 20 high resolution samples for the multi-fidelity approach. Training of single-fidelity VAE and compliance NN surrogate takes 0.17 and 0.09 seconds per iteration, respectively, whereas it takes 0.04 and 0.010 seconds per iteration for low-fidelity networks and 0.015 and 0.010 seconds per iteration for high-fidelity VAE and compliance networks respectively. All the training is done using NVIDIA GeForce RTX 3080. Thus, multi-fidelity approach gives significant improvement in the computational cost with similar or better performance than the single-fidelity approach. It is to be noted that, for a fair comparison between the performances of single- and multi-fidelity models, it is important to have similar total time which includes the time spent for data generation and model training. In the current experiment, the total computational time for multi-fidelity model is lower, and it has better performance compared to the single-fidelity model. As a future work, one can align the total training time for these models (by increasing the training data for multi-fidelity model) and draw a comparison between model performances at various computational budgets. 

\subsection{L-bracket with multiple loading uncertainty}

\begin{figure} 
\centering
\includegraphics[width=0.6\textwidth]{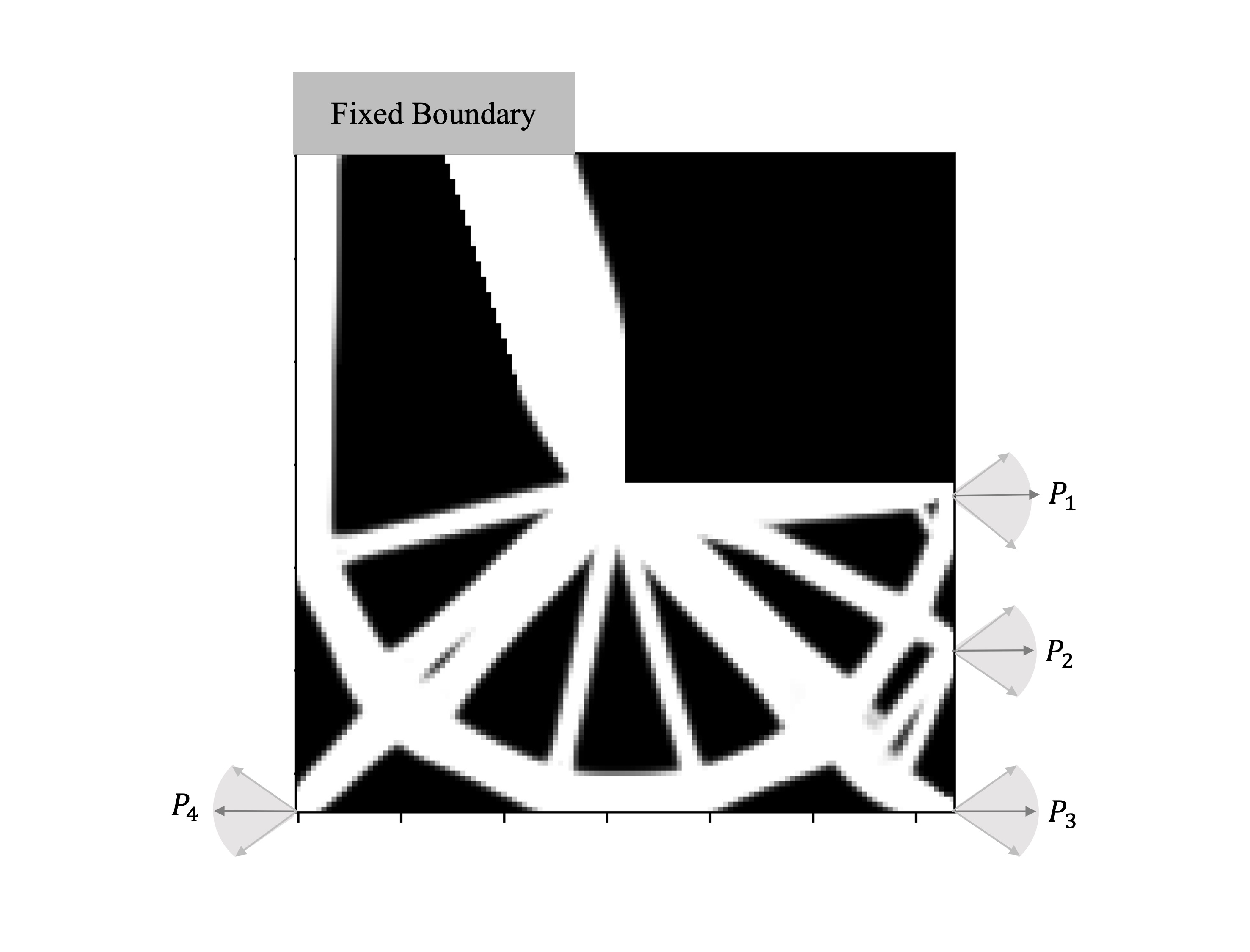}
\caption{An example of the structure used in our study. The top part of the L-shaped structure has a fixed boundary. The load is applied at 4 different points, $P_1$, $P_2$, $P_3$ and $P_4$, with uncertainty in all the four loading angles.}
\label{fig:multipoint}
\end{figure}

For the second example, we  consider the L-shaped structure with fixed boundary as the previous example, but with multiple point loads. In this case, we have considered 4 point loads at positions as shown in Fig.~\ref{fig:multipoint}. All the four loads have fixed magnitude, but uncertainty in the load angles. They are uniformly distributed between 0 and $\pi/2$. In this example, we consider the resolution of the images representing the topology to be $128\times128$. We consider higher resolution designs for this example, as more refined designs are required to capture the changes in design with the changes in multiple load angles. Here, the input \(\bm \theta\),  defining the geometry of the structure, is a \(n_{\bm \theta}=128 \times 128\) dimensional vector. The random input \(\bm \xi\) in this case are four random variables (i.e. \(n_{\xi}=4\)). The objective is to find the optimal design, \(\bm \theta^*\), via finding the optimal $\bm z^*$, which minimizes the robust compliance of the structure, \(Q_\text{rob}\), given the uncertain multiple load angles.

\begin{figure} 
\centering
\includegraphics[width=0.6\textwidth]{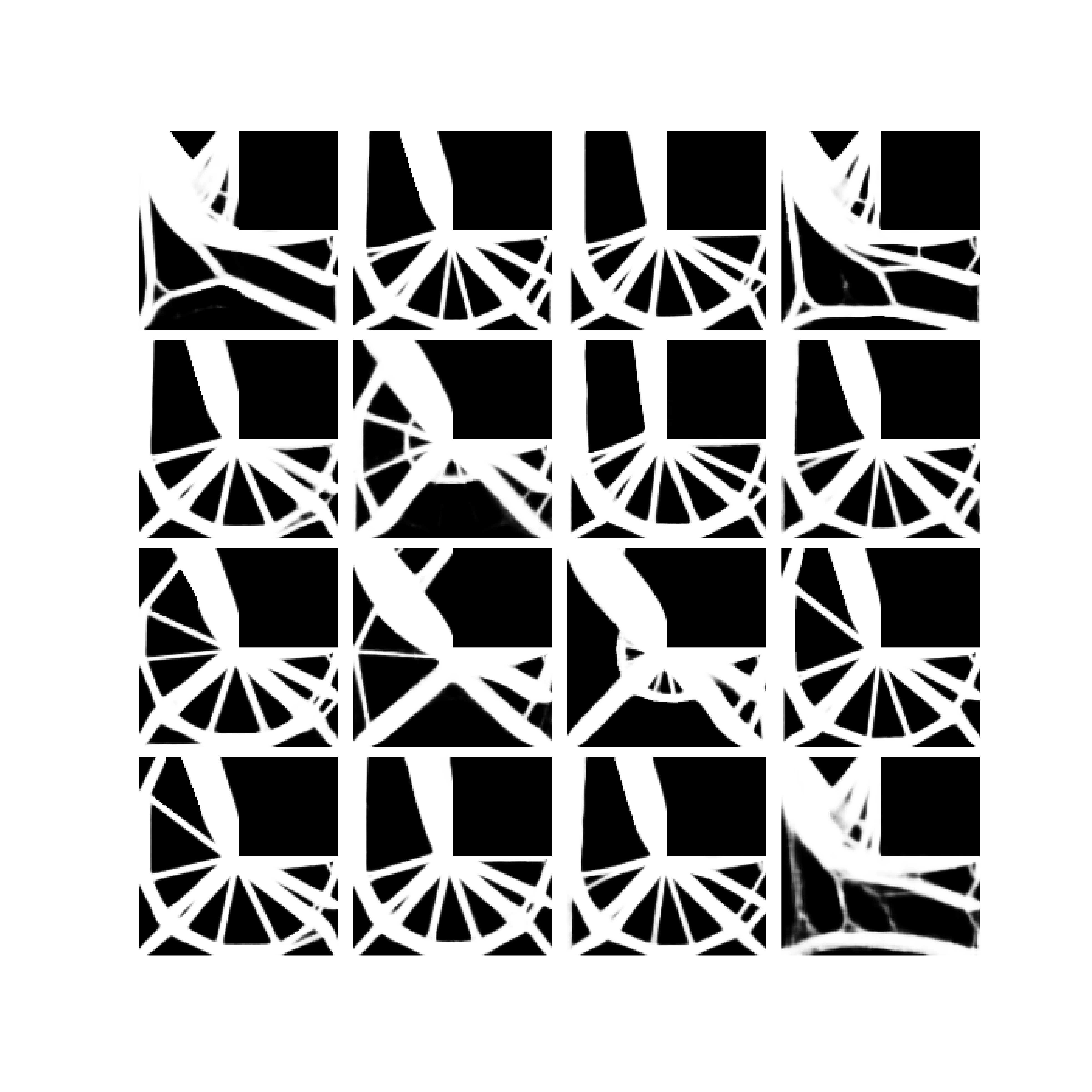}
\caption{A sample of images generated by sample z values with $|\bm{z}|=2$  from standard normal distribution for $n_\text{train}=600$. The sampled z values are fed into the the decoder part of the VAE to generate these images.}
\label{fig:fake_images_multi} 
\end{figure}

In order to solve the robust compliance minimization, the VAE is trained using a sample pool of 600 images of resolution $128 \times 128$. We use the same network architecture as the previous example with 7 hidden layers for both VAE and compliance NN surrogate. The input layer for both the networks, however, would have 16,384 ($128\times128$) nodes to match the dimensions of the geometries. The number of nodes in the first hidden layer is also increased to 7500 for better training stability. Fig.~\ref{fig:fake_images_multi} shows a set of candidate designs randomly generated by the decoder part of the trained VAE for L-bracket with multiple loads. 

\begin{figure} 
\centering
\subfigure[Best training sample, $Q_\text{rob}=388.01$]{\includegraphics[width=0.38\textwidth]{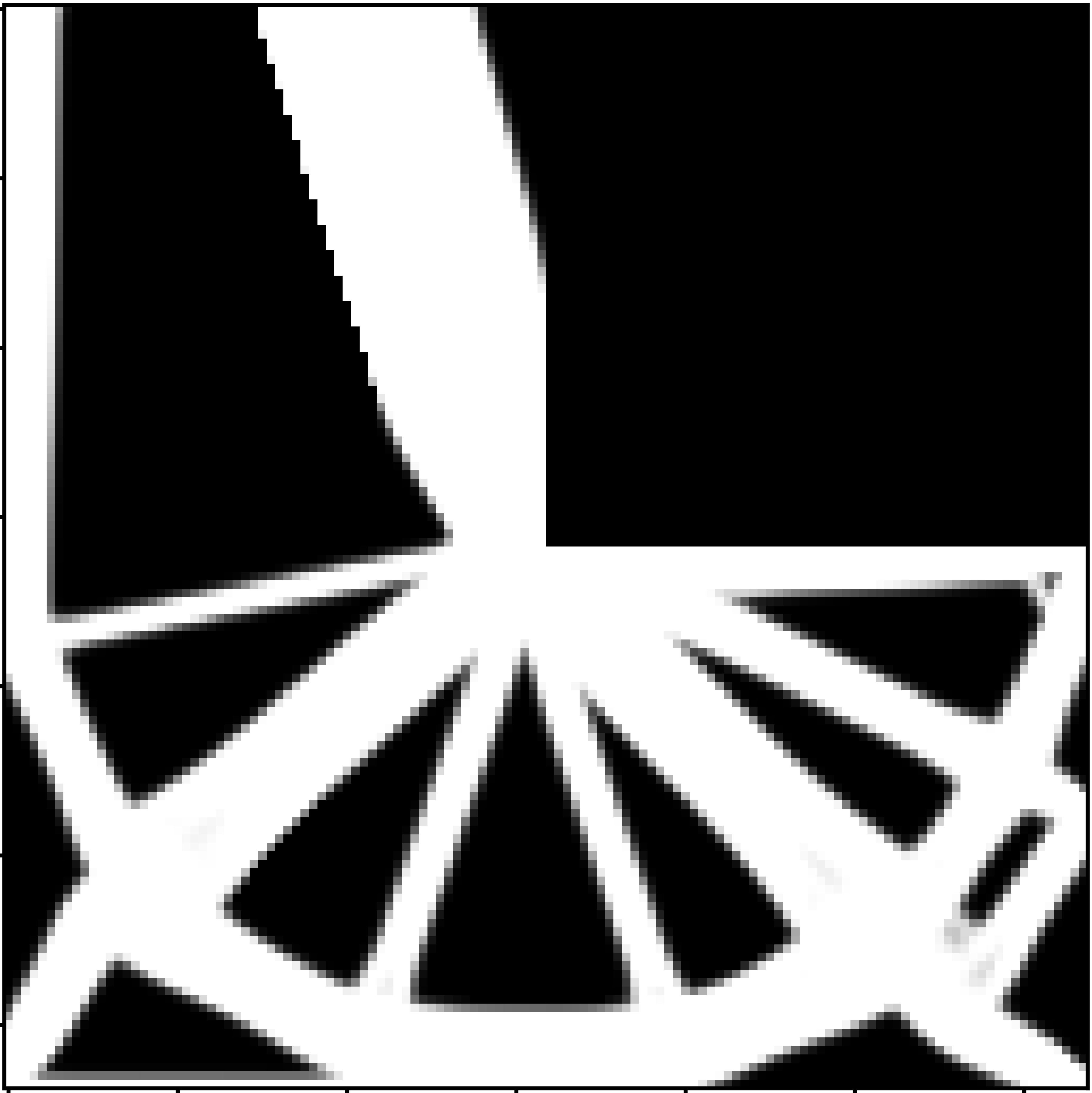}}\quad
\subfigure[Single-fidelity robust optimum ($n_\text{train}=600$), $Q_\text{rob}=382.10$]{\includegraphics[width=0.38\textwidth]{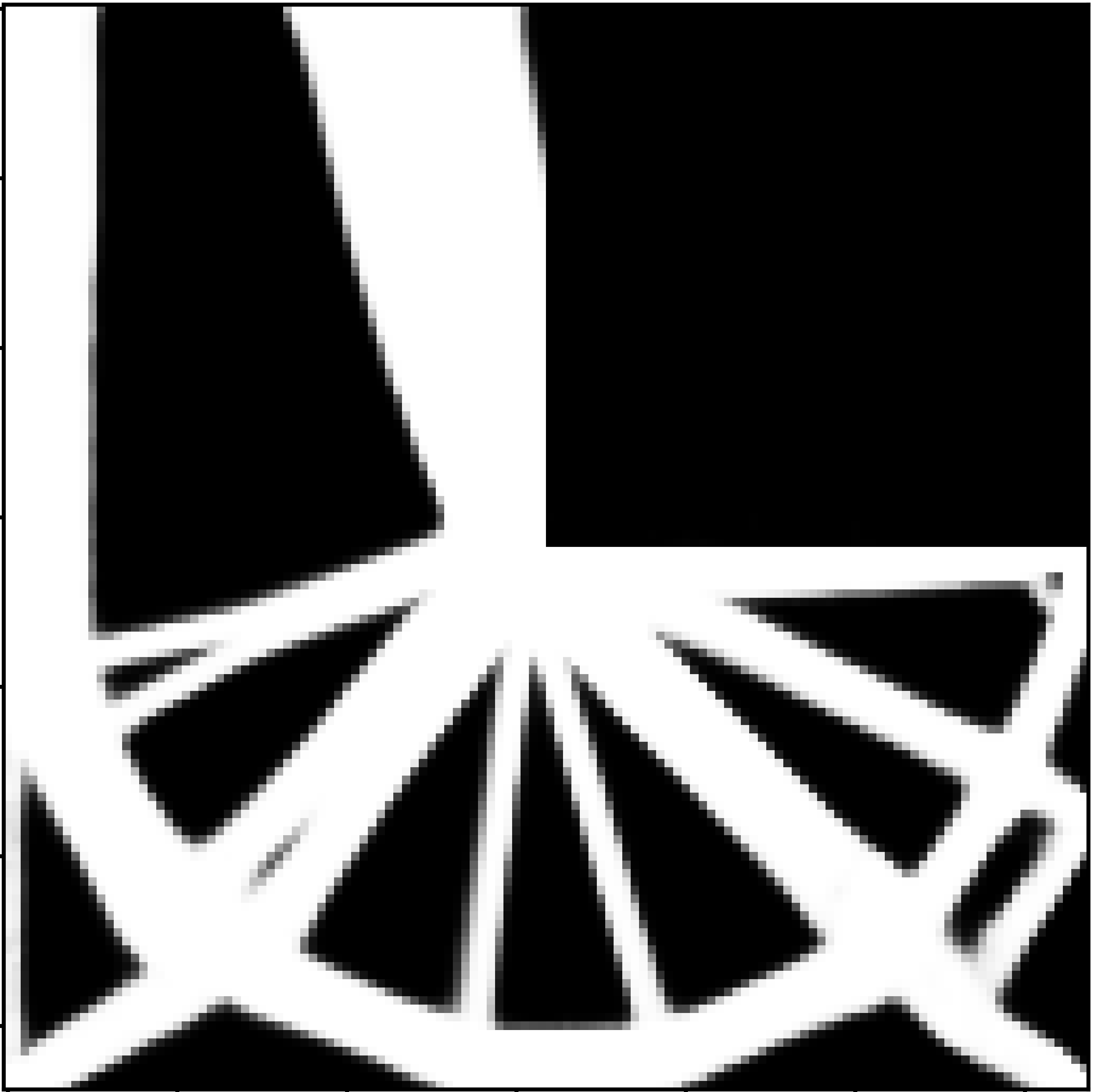}}\quad
\subfigure[Multi-fidelity robust optimum ($n_\text{train}=90$ for high-resolution, $n_\text{train}=700$ for low-resolution), $Q_\text{rob}=350.49$]{\includegraphics[width=0.38\textwidth]{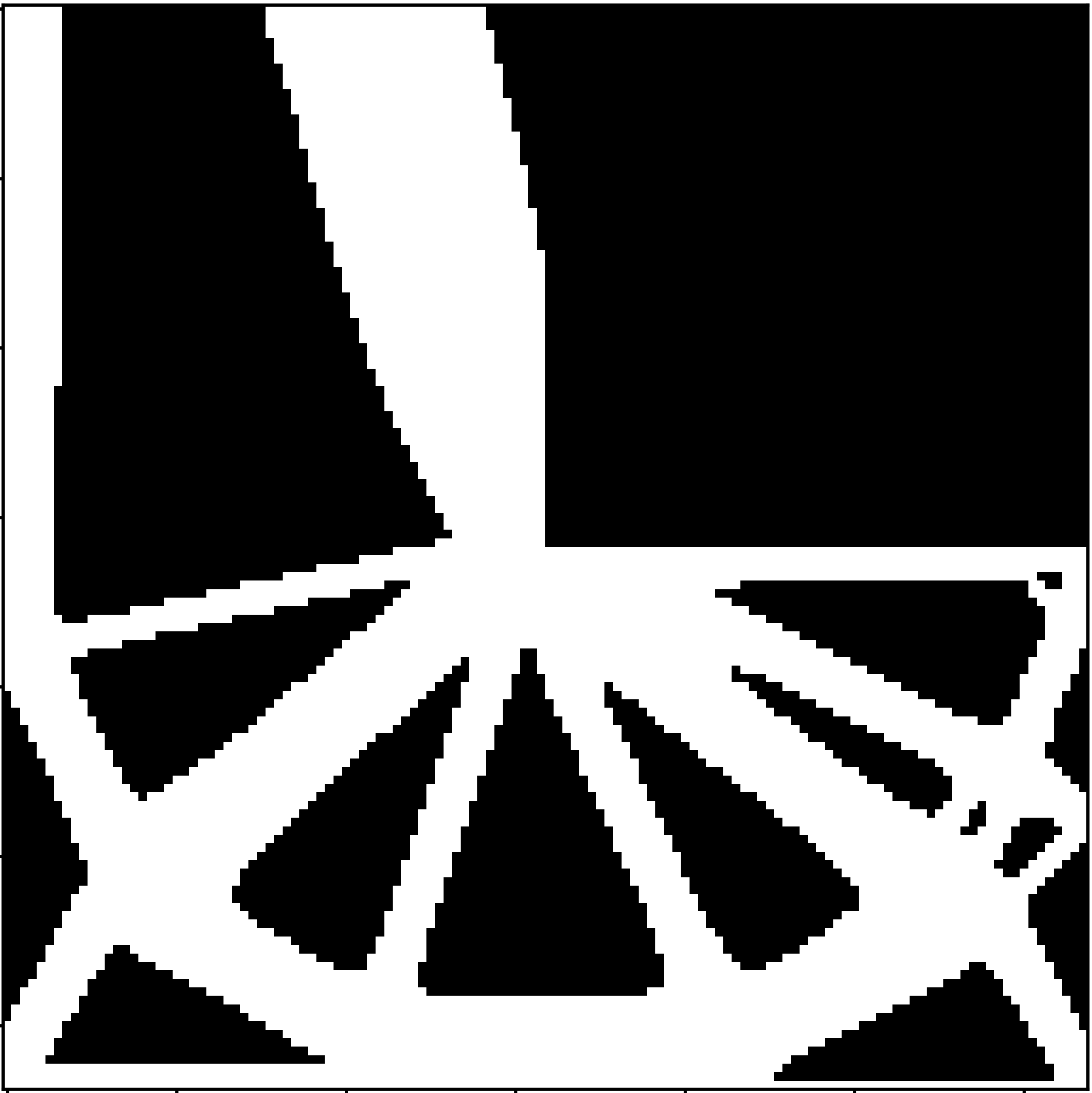}}\quad
\caption{The identified designs for the L bracket wth 4 point loads.}
\label{fig:multi_optimal} 
\end{figure}

Once the VAE and compliance neural networks are trained, similar to the previous example, we use gradient descent approach to find the optimal design for the multi-load case. Fig.~\ref{fig:multi_optimal} shows the comparison of the best design found in the training set and the optimal design found via gradient descent algorithm. It can be observed that gradient descent gives a better design, in terms of robust compliance, compared to the samples in the training data. This is in accordance with our observations from the previous example.

\begin{figure} 
\centering
\includegraphics[width=0.6\textwidth]{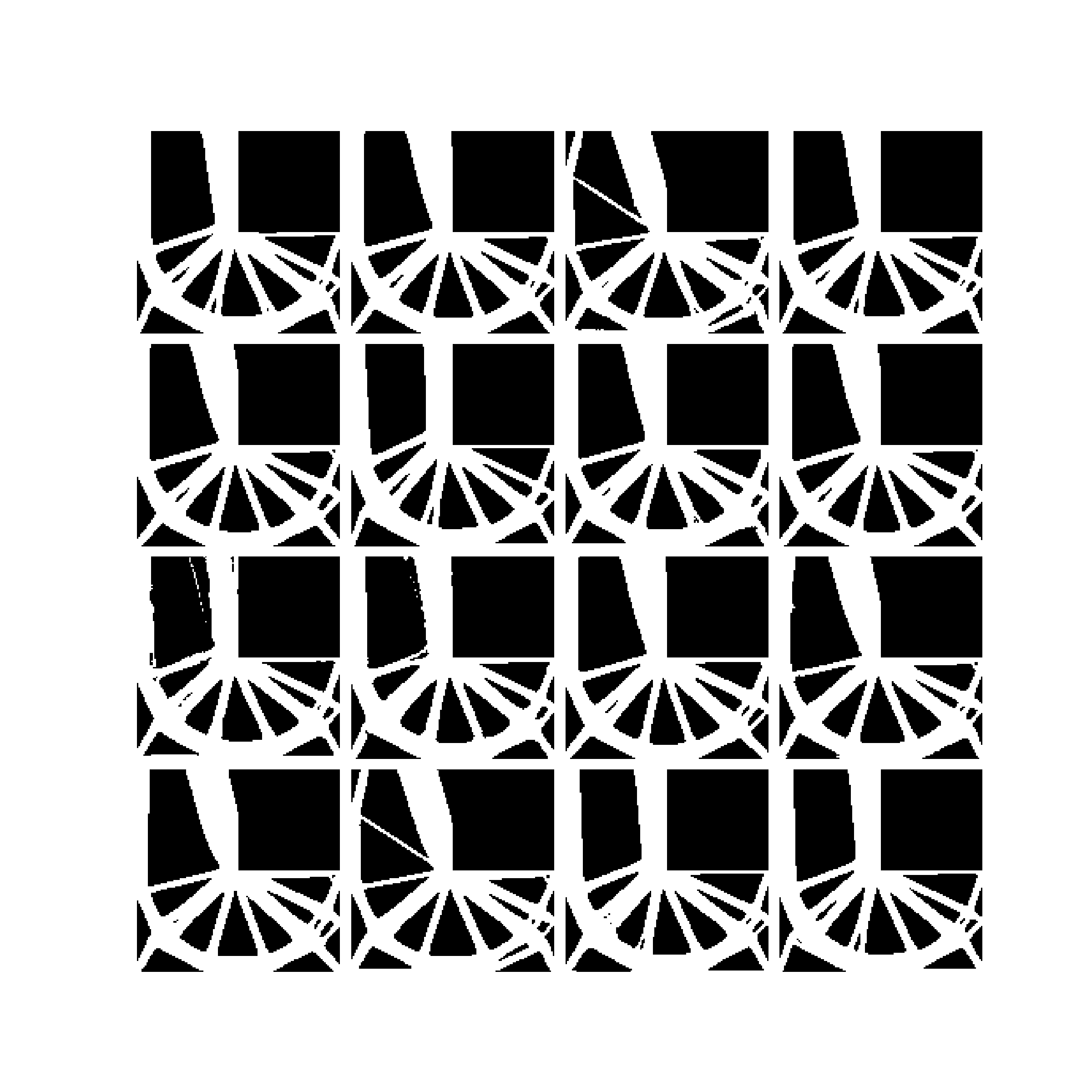}
\caption{A sample of images generated by sample z values with $|\bm z|=2$  from standard normal distribution for high-fidelity VAE for $n_{train}=90$. The sampled z values are fed into the the decoder part of the VAE to generate these images.}
\label{fig:fake_images_multi_hf} 
\end{figure}

To create the multi-fidelity framework, the pre-trained network is the same low-fidelity network of the previous section, i.e., the one trained with low-resolution images of dimension $30\times30$, with single load, instead of multiple loads. The high-fidelity network is trained with 90 high-resolution images with  $128\times128$ resolution with multiple loading. Our goal is to verify if high-fidelity VAE can still learn to generate topology variability using a pre-trained network trained using topologies belonging to a different distribution. Fig.~\ref{fig:fake_images_multi_hf} shows the designs randomly generated by the decoder part of the high-fidelity VAE. While we do not see as much variability in the randomly generated images as in Fig.~\ref{fig:fake_images_multi} produced by VAE trained with 600 images (from here on referred to as single-fidelity VAE), upon close observation, we can see some features of the generated images that the model has learned from the low-fidelity VAE, like the extra elements in the top half of some of the images and the curvature of the bracket near the fixed boundary location in some, which are not present in the 90 images used for the training.

The trained multi-fidelity network is then used for finding the robust optimal design via the gradient descent optimization. Fig.~\ref{fig:multi_optimal}(c) shows the optimal design obtained through the multi-fidelity approach and we can observe that, the identified optimal design has a robust compliance of 350.49, which is much lower than that of single-fidelity network (382.10) and the one found in the training data (388.01). Moreover, it takes 12.75 hours to generate 600 multi-load designs of resolution $128\times128$ for the training of single-fidelity networks, whereas it takes only 1.9 hours to produce 90 designs needed for the high-resolution networks in the multi-fidelity approach. Also, it takes 1.2 seconds per iteration for single-fidelity VAE and compliance training whereas it takes 0.18 seconds per iteration for high-fidelity network training using NVIDIA GeForce RTX 3080. Thus, we can conclude that the multi-fidelity approach saves significant computational cost compared to single-fidelity network and provides good improvement in the model performance.

\section{Conclusions}

Input uncertainties and computational challenges inherent in topology optimization approaches using finite element solvers motivate the utilization of deep learning algorithms to replace the bottle neck steps of the optimization process. In this paper, we studied how deep learning based techniques and neural networks can be used for an efficient parametrization of design topologies  and also to provide an approximate forward model for the evaluation of robust compliance that can be easily differentiated and used in a gradient-based optimization. Using two examples, we highlighted how the proposed approach can offer computational efficiency with high accuracy, which can be further enhanced by incorporating a multi-fidelity framework on the network architecture and training. An important advantage of using deep learning in shape parametrization is its capability for automatic feature engineering/detection, using minimal amount of training data. This removes the need to explicitly impose the design constraints, such as volume fraction, in the loss function. 

In this work, we used VAEs to turn the high dimensional  optimization problem into a low dimensional one. Even though VAE was able to produce better designs than the training data, we could not conclude that it can produce a robust optimal design that is better than what a finite element approach offers. The reason is that the quality of our optimal design is  dependent on the search subspace, informed by the training data, which are the designs obtained from the deterministic optimization problem with different loading realizations. It should be noted that if the actual robust design has topology features that lie  outside the search subspace, our gradient based approach may have limitations in reaching that specific topology. Therefore, as a next step, we are exploring the use of sub-optimal designs from SIMP for the training and evaluating if they can produce a design closer to the robust optimal design given the design constraints. This also points at the need for lower reliance on the training data for design generation. To this end, we are studying how we can replace a data driven neural network with a physics informed neural network, where design generation would be based on the governing physical equations, along with minimal training data. 
The other extensions to this work to be addressed in the future studies are (1) evaluating the structural integrity of the optimal design from the gradient descent process; and (2) incorporating the design constraints of the optimization problem in the neural network architecture.

\bibliographystyle{unsrt}  
\bibliography{References}  

\end{document}